\documentclass[runningheads]{llncs}
\usepackage{graphicx}
\usepackage[misc]{ifsym}

\usepackage{graphicx}
\usepackage{adjustbox}
\usepackage{todonotes}
\usepackage{color,soul}
\usepackage{tablefootnote}
\usepackage{multirow}
\usepackage{amsmath}
\usepackage{amsfonts}
\usepackage{array}

\usepackage{color,soul}
\usepackage{booktabs}
\usepackage{hyperref}

\newcolumntype{C}[1]{>{\centering\arraybackslash}m{#1}}

\usepackage{geometry}
\newif\ifrenderfigures
\renderfigurestrue
\usepackage[square,numbers]{natbib}
\PassOptionsToPackage{numbers, compress}{natbib}
\begin{document}
\title{Ablation Based Counterfactuals}
%
%
\author{Zheng Dai\inst{1}(\Letter)\orcidID{0000-0002-8828-1075} \and
David K Gifford\inst{1}(\Letter)\orcidID{0000-0003-1709-4034}}
\authorrunning{Z. Dai et al.}
%
\institute{Computer Science and Artificial Intelligence Laboratory, Massachusetts Institute of Technology, Cambridge MA 02139, USA \email{\{zhengdai,gifford\}@mit.edu}}
\maketitle              
\begin{abstract}
Diffusion models are a class of generative models that generate high-quality samples, but at present it is difficult to characterize how they depend upon their training data.  This difficulty raises scientific and regulatory questions, and is a consequence of the complexity of diffusion models and their sampling process. To analyze this dependence, we introduce \emph{Ablation Based Counterfactuals} (ABC), a method of performing counterfactual analysis that relies on model ablation rather than model retraining. In our approach, we train independent components of a model on different but overlapping splits of a training set. These components are then combined into a single model from which the causal influence of any training sample can be removed by ablating a combination of model components. We demonstrate how we can construct a model like this using an ensemble of diffusion models. We then use this model to study the limits of training data attribution by enumerating full counterfactual landscapes, and show that single source attributability diminishes with increasing training data size. Finally, we demonstrate the existence of \emph{unattributable samples}.

\end{abstract}
\section{Introduction}

Diffusion models have emerged as powerful tools for modeling and sampling from complex natural distributions, and have achieved remarkable results in a wide array of applications ranging from text conditioned image generation \cite{ramesh2022hierarchical}, video generation \cite{ho2022video}, audio synthesis \cite{zhang2023survey}, and even therapeutic design \cite{luo2022antigen}. To attain these performances, these models often need to be trained on massive corpuses of training data. The amount of training data, along with the complexity of the models and their sampling process, makes it difficult to assess how training data ultimately impacts the final generated sample. 

In this work, we introduce the concept of an Ablation Based Counterfactual (ABC), where a model is trained in a way such that the causal influence of a given piece of data can be surgically removed by ablating select pieces of the model, forgoing the need for retraining. This enables us, for the first time, to generate entire leave-one-out counterfactual landscapes without the use of approximate methods. By enumerating these landscapes, we show that the ability to attribute a generated sample to a piece of training data deteriorates as training set sizes increase, culminating in the phenomenon of \emph{unattributable samples}. 
An unattributable sample is a sample that cannot be attributed to any single piece of training data. The existence of such samples have serious scientific and policy implications which we touch on.

\subsection{Related work}

This work is a refinement and significant extension of our previous work~\cite{dai2023training}.

Diffusion models were originally introduced to machine learning by \citet{sohl2015deep}. In our work, we use the model architecture and training process from \citet{ho2020denoising}, which improved upon and popularized the original diffusion model. We also use latent diffusion for more complex sample generation tasks, which is a technique introduced by \citet{Rombach_2022_CVPR}.

Assessing the impact of training data on models is an active area of research with wide applicability, including model interpretability \cite{koh2017understanding}, machine unlearning \cite{nguyen2022survey}, data poisoning \cite{chen2017targeted}, fairness \cite{mehrabi2021survey}, and privacy \cite{shokri2017membership}. Assessing the impact of training data typically relies on a retraining based paradigm, where a new counterfactual model is trained with an incomplete training set. The new model is then compared with the original model to assess the impace of the missing training data. Computing a retrained model is very expensive, and existing methods seek to circumvent retraining through approximate methods~\cite{koh2017understanding, pruthi2020estimating, park2023trak}. The true interpretations of these approximations are sometimes disputed \cite{basu2020influence}. Our method by contrast establishes an ablation based paradigm, where counterfactuals are assessed via model ablation, allowing the exact assessment of the counterfactual scenario without retraining in contrast to a retraining based paradigm.

In this work we use ensembles of diffusion models as a means of creating ABC models with sufficient redundancy to allow ablation. Ensembles of diffusion models was previously seen in~\cite{balaji2022ediffi}, although the ensemble in their work operated in sequence, so each denoising step is still taken by a single model, making it impossible to ablate ensemble members while preserving functionality.

\subsection{Our Contributions}

Our contributions are the following:
\begin{enumerate}
    \item We present ablation as an alternative to the retraining based paradigm for creating counterfactual scenarios. Instead of retraining the model without pieces of the training data, we train different pieces of the model on different parts of the data. We can then remove the causal influence of given pieces of training data by surgically ablating select pieces of the model.
    \item We show how such a model can be created using an ensemble of diffusion models. We demonstrate the viability of the diffusion ensemble for the task of image generation.
    \item We show that the attributability of diffusion models decreases with increasing training set size. Furthermore, we demonstrate the existence of samples that are \emph{unattributable}, meaning that they cannot be attributed to any single source of training data.
\end{enumerate}
\section{Methods}

\subsection{Preliminaries}

Training data (e.g. creative works) are derived from data sources (e.g. authors). Our aim is to study the impact a data source has on a final generated sample. The study of the impact of individual points of data is a special case of this where each data point is its own individual data source.

There are two schools of analyzing this, which we term \emph{visual} analysis and \emph{counterfactual} analysis.
In \emph{visual analysis}, the impact of a data source is inferred by the visual similarity of its training data to the generated sample. This can be effective, for example when the generative model directly copies the training data~\cite{somepalli2023diffusion}. It is often implicitly used as a sanity check for other forms of analysis. However we show later in Section \ref{sec-res-viscf} that this form of analysis can be misleading.

In \emph{counterfactual analysis}, the question ``what if this data source did not exist?'' is considered. To operationalize this, the counterfactual scenario is often computed via retraining or some approximation thereof, and the outcome of the counterfactual scenario is compared with the true scenario. In this work we are primarily concerned with performing counterfactual analysis. Specifically our goal is to compute counterfactual samples.  A \emph{counterfactual sample} is one generated in the counterfactual scenario instead of the factual sample generated in the factual scenario. Significant differences in factual and counterfactual samples are indicative of influential data sources.

Each counterfactual sample, in addition to being associated with a factual sample, is also associated with the data source that was missing in the counterfactual scenario. For each factual scenario, the number of counterfactual scenarios are commensurate with the number of data sources. We will refer to the set of all possible counterfactual samples for a given factual sample as its \emph{counterfactual landscape}. A limitation of the present analysis is that we do not consider counterfactual scenarios where multiple data sources are combinatorially removed.

\subsection{Ablation based counterfactuals}

A sample is generated through the following chain of events: data sources $s_i$ create training data $x_i$. Training data $x_i$ is used to train model parameters $\Theta$. Model parameters $\Theta$ are used, along with sampled noise $\varepsilon$ (which we will refer to as \emph{exogenous noise}) to generate a sample $y$. This is illustrated on the left panel of Figure \ref{fig1}.

\ifrenderfigures
\begin{figure}
  \begin{center}
  \includegraphics[width=\linewidth]{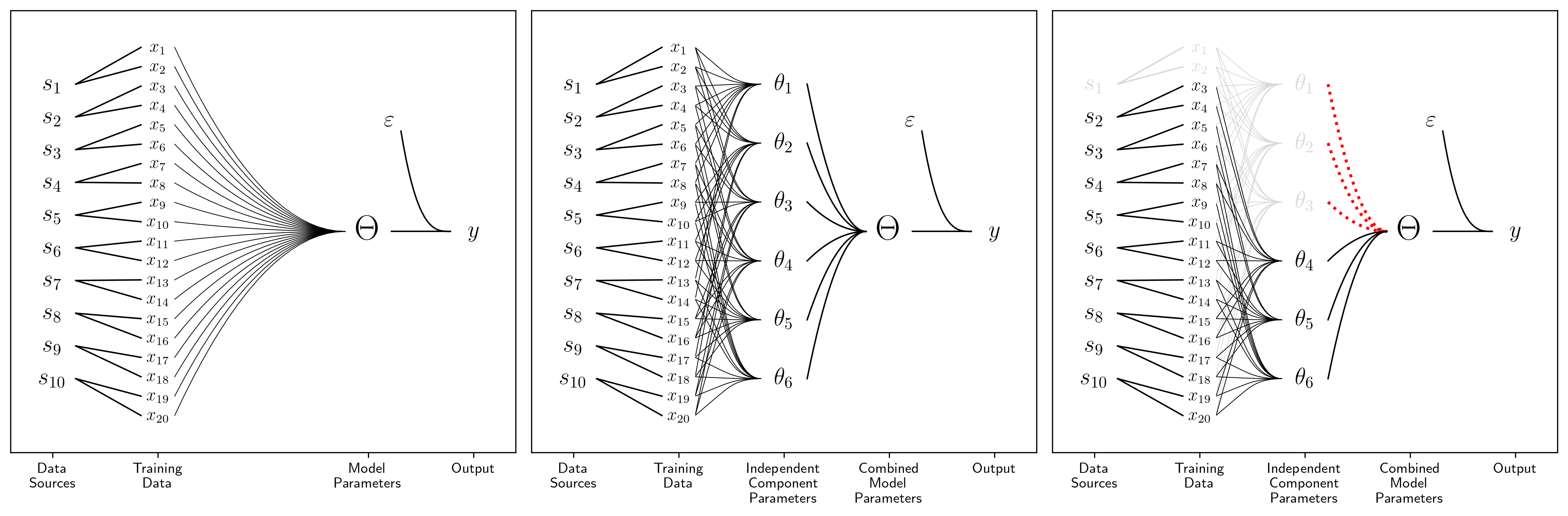}
  \end{center}
  \caption{Causal flow from data sources to generated sample}
  Causality flows from left to right in all panels. In the right panel, the causal link from $\theta_1$, $\theta_2$, and $\theta_3$ to $\Theta$ is broken (illustrated by the broken red line). The sections of the causal graph that are no longer connected to the output $y$ are rendered in a lighter color. Note that thanks to our causal setup, the only data source that has its connection to $y$ removed is $s_1$. Close inspection would reveal that for any $s_i$, it is possible to remove a combination of $\theta_j$ such that $s_i$ and only $s_i$'s connection to $y$ is severed.
  \label{fig1}
\end{figure}
\fi

We wish to study the counterfactual. In the above setup, the only way to fully disentangle the effect of a data source and assess the counterfactual is to retrain the model parameters from scratch. We then pump the same exogenous noise through the retrained model to obtain a counterfactual sample. We refer to such a counterfactual as a retraining based counterfactual (RBC).

The main problem with the RBC approach is the computation required to retrain the model, which makes it infeasible if we wish to assess a large quantity of counterfactuals. We propose an ablation based approach as an alternative.

We first reorganize our causal structure. Instead of making the model parameters the causal result of the entire dataset, we split the parameters $\Theta$ into multiple small components $\theta_i$. The components must be redundant (i.e. there are no critical components that the model cannot function without). Each of these components are then trained on a subset of the training data. This change in the causal structure is illustrated in the middle panel of Figure \ref{fig1}.

The key insight is that to break the causal link between a piece of training data $x_i$ and the generated sample $y$, \emph{it is sufficient to remove all the components $\theta_i$ that have been trained on it}, circumventing any need to retrain. If organized correctly, the causal link between any data source and the generated sample can be broken without breaking the causal link between any other training data  and the generated sample (see Section \ref{sec-meth-encoded}).

Therefore, given a generated sample $y$ and the exogenous noise $\varepsilon$ used to generate it, we can generate a counterfactual sample that removes all influence from a data source $s_i$ by ablating all components of the model that has been trained on data produced by that data source and regenerating the sample. We will term such samples as \emph{Ablation Based Counterfactuals}, or ABCs for short.

\subsection{Diffusion ensembles enable ablation}
\label{sec-meth-encoded}

Ensembles of models are amenable to ablation, where each model of the ensemble has an identical architecture and are trained on different splits of the training data. We find this offers sufficient redundancy to permit ablation of ensemble members.

To generate our ensemble, we first extract various training splits from our training set. We then independently train a diffusion model on each split. A common convention is to consider a diffusion model to be a deep learning model (oftentimes a U-Net) that takes as input the linear combination of an image and some noise and predicts the what the added noise was~\cite{ho2020denoising}. Our ensemble does exactly that, except instead of a U-Net it's an ensemble of U-Nets. The input is passed to each model, each of which produces a prediction of the added noise. The arithmetic mean of the predicted noise is then taken to produce the output of the ensemble. If component models of the ensemble are ablated, we take the arithmetic mean of the outputs of the remaining ensemble members. The viability of this approach is demonstrated in Section \ref{sec-ref-viability}.

\newcommand{\bcode}{\mathcal{C}}

The remaining consideration is how to determine the training splits the component models are trained with. To determine the training splits we first pick a binary code $\bcode$ that contains at least $N$ codewords of length $n$, where $N$ is the number of data sources. Each data source is then assigned a unique codeword from $\bcode$. We then create $n$ training splits, where the $i$th training split will contain a piece of training data if and only if the code assigned to its data source has a 1 in position $i$. To ensure that each data source is represented equally within the ensemble, we require each codeword of $\bcode$ to have equal Hamming weight. This is sufficient to ensure the following:

\begin{theorem}
\label{thm-1}
    Given a data source $s$, define $S(s)$ as the set of models in the ensemble whose training split contains data produced by $s$. Then the set difference $S(s)\setminus S(s')$ is not the empty set for any distinct data sources $s$ and $s'$.
\end{theorem}

Therefore, if we ablate away all and only those models that have been trained on a given data source, the influence of no other data source is removed.
The proof is provided in Appendix \ref{appdx-proofs-thm1}.

We can also show that given $N$ data sources, an ensemble with $\mathcal{O}(log(N))$ members is sufficient to account for all of them (see Appendix \ref{appdx-proofs-sperner}), representing an exponential time save for model training when compared to the retraining paradigm. A full runtime analysis is provided in Appendix \ref{appdx-proofs-runtime}.

\subsection{Differential ablation enable efficient approximation of ablation}
\label{sec-differentialAblation}

While the ABC paradigm removes the bottleneck of retraining models, sample generation can still be expensive, especially in diffusion models. To make the computation of ABCs efficient across very large datasets, we introduce \emph{differential ablation} as a method of approximating ABCs.

Let $f_1, ..., f_n$ denote the $n$ diffusion models that make up the ensemble. Let $c \in \mathbb{R}^n$, and define $f\cdot c$ as a function where for any input $x$ we have $(f\cdot c)(x)= \sum_{i=1}^nf_i(x)\frac{c_i}{n}$. When $c$ is the vector where each entry is 1, $f\cdot c$ is just the normal ensemble. If $c$ is a vector where half the entries are 0 and the other half are 2s, then $f\cdot c$ is an ensemble where $f_i$ is ablated if $c_i$ is 0.

Given some exogenous noise $\varepsilon$ and a diffusion model $g$, let $\mathcal{G}(g, \varepsilon)$ be the output of the diffusion process where $g$ is used as the model and $\varepsilon$ is used as the exogenous noise. We then consider the Taylor expansion of $\mathcal{G}(f\cdot c, \varepsilon)$ around the all 1 vector (denoted $u$):

\begin{equation}
    \mathcal{G}(f\cdot c, \varepsilon) = \mathcal{G}(f\cdot u, \varepsilon) + \frac{\partial \mathcal{G}(f\cdot x, \varepsilon)}{\partial x}\bigg|_{x = u} (c - u) + \mathcal{O}(\|c - u\|^2)
\end{equation}

Where $\frac{\partial \mathcal{G}(x, \varepsilon)}{\partial x}$ is the Jacobian, which is an $m$-by-$n$ matrix where $m$ is the dimension of the generated sample. This can be computed efficiently by performing $n$ rounds of forward mode automatic differentiation. After that, approximating $\mathcal{G}(f\cdot c, \varepsilon)$ can be done by taking the matrix-vector product between the Jacobian and $(c - u)$ and truncating the remainder of the Taylor expansion. As mentioned earlier, by setting $c$ appropriately $f\cdot c$ will become an ablated ensemble, so $\mathcal{G}(f\cdot c, \varepsilon)$ will take on the value of the corresponding ABC. We call this approximation method \emph{differential ablation}.

Differential ablation is able to accurately predict the effects of ablation, which we demonstrate in Appendix \ref{appdx-res-diffabl}.
\section{Results}

Our first goal is to establish the viability of our methods. First, we demonstrate that an ensemble of diffusion models is a viable generative model. We then compare our ablation based paradigm to the retraining based paradigm for analyzing counterfactuals, and show that they exhibit similar patterns.

We then evaluate the use of ABCs to detect influential data sources, and find that it produces visually reasonable results when training sets are small, but produce less visually reasonable results when training sets become larger. Conversely, we find that visual similarity between a generated sample and a training sample does not necessarily mean that the training sample was influential in the generation of the generated sample, especially when training sets are large. These two observations establishes a discrepancy in visual and counterfactual analysis at large training set sizes.

Finally, we show that in general the influence of individual data sources begins to vanish as training sets get large, culminating in models that generate unattributable or nearly unattributable samples.

\subsection{Diffusion ensembles are viable generative models}
\label{sec-ref-viability}

We train a total of 23 diffusion ensembles, which are summarized in Table \ref{table0}. Models are trained similarly to how they were trained by~\citet{ho2020denoising} (details are provided in Appendix \ref{appdx-models-training}). The ensembles are named by the subsets of their publicly accessible training datasets (see Appendix \ref{appdx-exp-data-prep}), and are denoted MNIST\_256, MNIST\_512, MNIST\_1024, MNIST\_2048, MNIST\_4096, MNIST\_8192, FASHION\_256, FASHION\_512, FASHION\_1024, FASHION\_2048, FASHION\_4096, FASHION\_8192, CIFAR-10\_500, CIFAR-10\_50000, CIFAR-100\_500, CIFAR-100\_50000, CelebA\_BA\_512, CelebA\_BA\_8192, CelebA\_SD\_512, CelebA\_SD\_8192, MetFaces, Landscapes, and ArtBench. Each ensemble consists of 22 members, with the exception of MetFaces and Landscapes, which contain 24 members. Conventionally, the name denotes the public dataset, and the number denotes the size of the counterfactual landscape. BA and SD refer to different autoencoders used to encode images to latent space, with BA being a basic autoencoder trained in-house (see Appendix \ref{appdx-otherModels-ba}) and SD being the Stable Diffusion autoencoder. A summary of datasets and ensembles can be found in Appendix \ref{appdx-experimental-setup}.

Select samples generated by the ensembles are shown in Figure \ref{fig2}. In addition to the ensemble, for each ensemble we train a single diffusion model as a control. We calculate the Frechet Inception Distance (FID)~\cite{heusel2017gans} based on 10240 samples generated from each ensemble and control model, and find that they perform comparably.

\ifrenderfigures
\begin{figure}
  \begin{center}
  \includegraphics[width=0.46\linewidth]{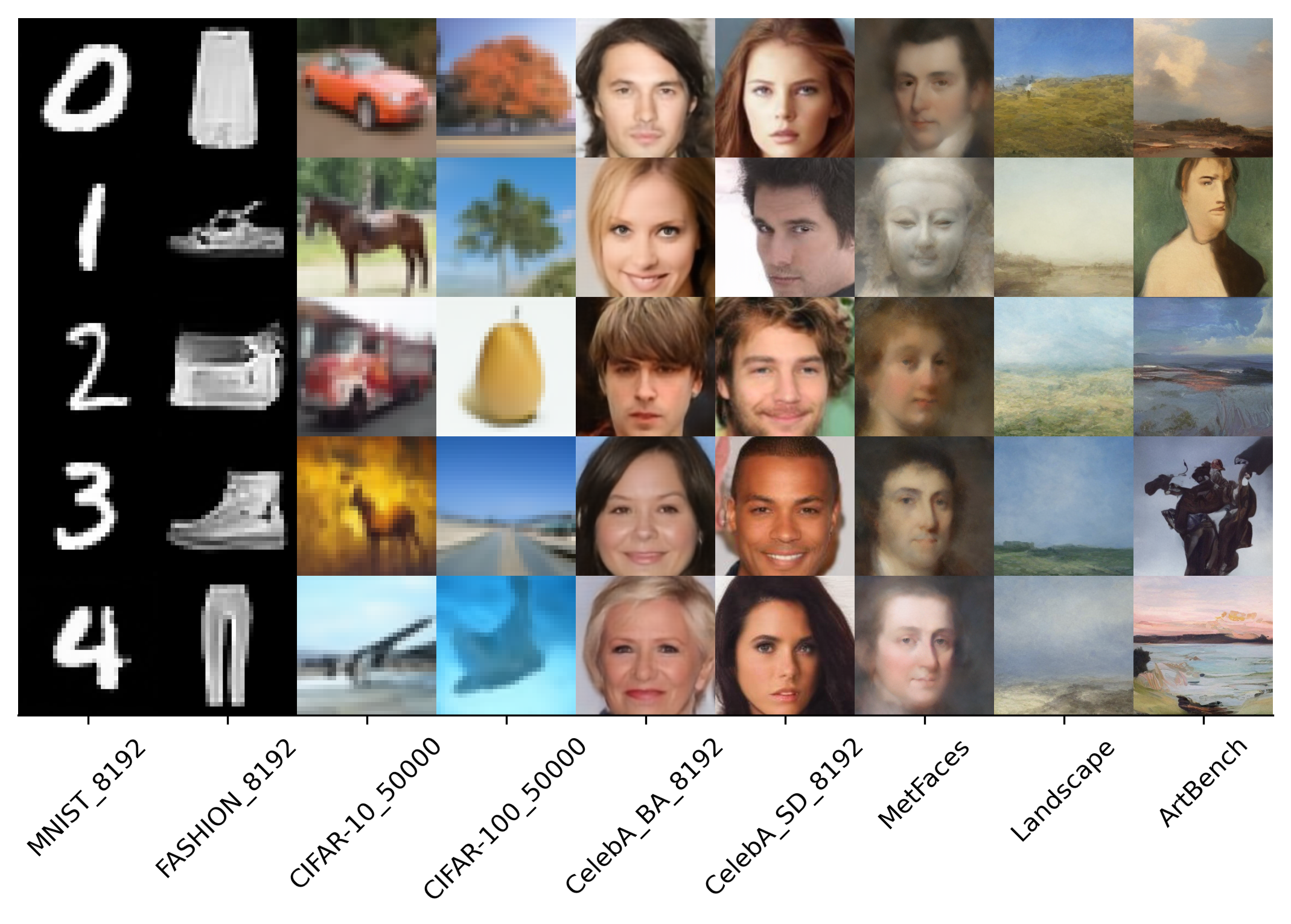}
  \includegraphics[width=0.48\linewidth]{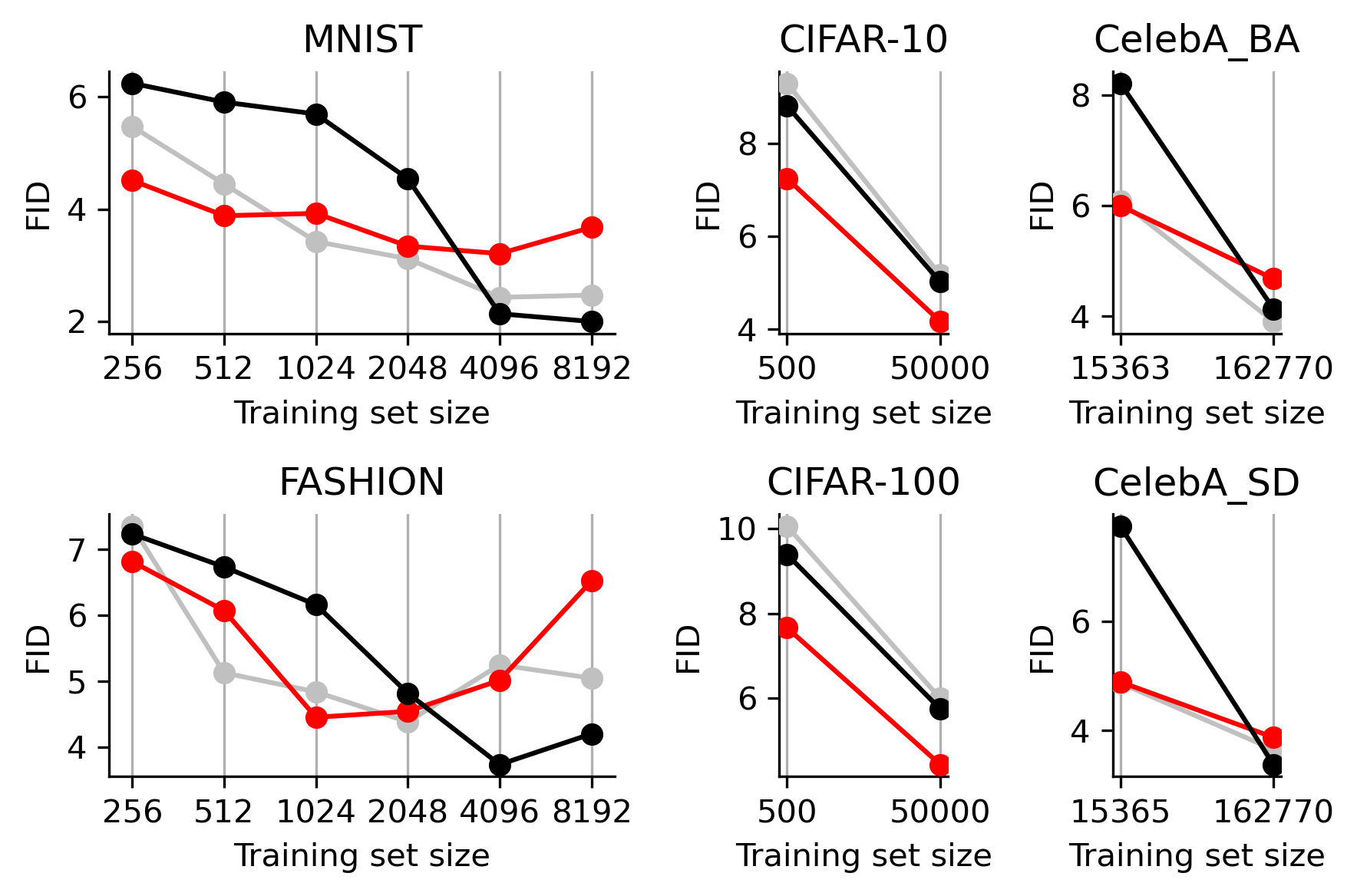}
  \end{center}
  \caption{Ensembles of diffusion models generate good quality samples}
  We show select samples generated from diffusion ensembles in the left panel. We trained ensembles on differently sized subsets of given datasets. We plot their FIDs as functions of training set sizes on the right in black. In red we plot the FIDs attained by single models that were trained on the entire dataset, and in grey we plot the FIDs of a single member of the ensemble when used as its own model. While ensembling appears to hurt performance at low training set sizes, it often boosts performance at larger training set sizes. Additional discussion on FID appears in Appendix \ref{appdx-res-fid}.
  \label{fig2}
\end{figure}
\fi

\subsection{Ablation based counterfactuals are comparable with retraining based counterfactuals}
\label{sec-res-benchmark}

To benchmark ABCs, we compare ABCs (counterfactuals generated through ablation) against RBCs (counterfactuals generated by retraining).
First, we train ensembles of size 6 on MNIST and FASHION, where each image class is considered to be a data source. We then train 11 models: one model on the full dataset and 10 models, each one on the full dataset minus a single image class. We then compare the ensembles' and models' ability to generate members of different classes in the full ensemble/model to that of the ablated ensembles/retrained models. The classes are determined using a ResNet-18~\cite{he2016deep} (see Appendix \ref{appdx-otherModels-mnist}). We find that the distribution of classes and their diminishment in the ablated/retrained models to be comparable in Figure \ref{fig3}.

\ifrenderfigures
\begin{figure}
  \begin{center}
  \includegraphics[width=0.9\linewidth]{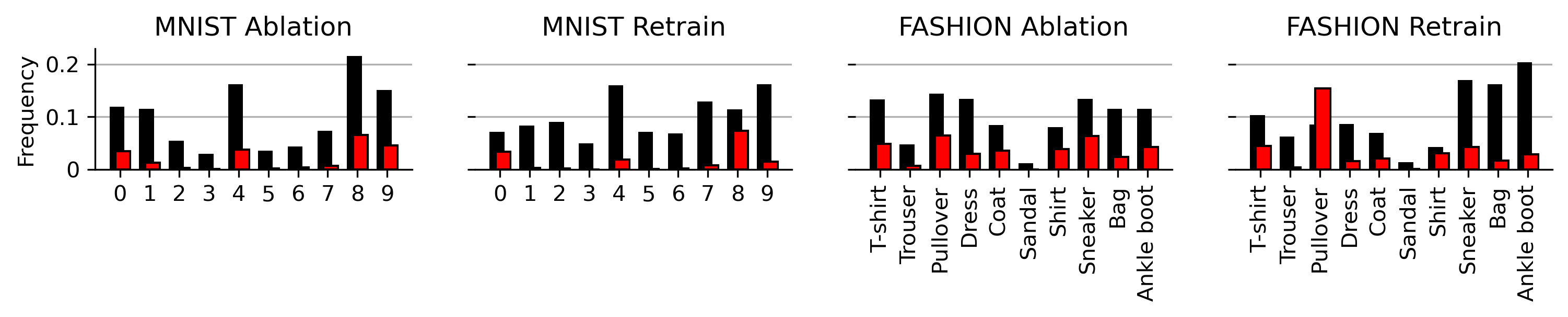}
  \end{center}
  \caption{Comparing ABCs (ablation) to RBCs (retraining)}
  The four panels show the frequencies that images of various MNIST/FASHION classes are generated at as black bars by ensembles/models trained on the full dataset. The red bars indicate the adjusted frequency when the class is removed, either through ablation or retraining. Frequencies are calculated from 5100 samples.
  The distributions and amount of diminishment are comparable between ablation and retraining for both MNIST and FASHION. We note that the frequency of pullovers appears to rise upon removing all pullovers in the training set and retraining: this is likely due to misclassification, since pullovers, dresses, and shirts all look alike at a 28x28 resolution.
  \label{fig3}
\end{figure}
\fi

We then train ensembles of size 16 on subsets of MNIST and FASHION of size 384, where each image is its own data source. We then trained 385 models: one model on the full subset, and one for each leave-one-out subset. To maximize consistency, the initial parameters and the order and contents of the minibatches were kept consistent for the training of each model, and the removed image was replaced with an uninformative all black image rather than removed.

Since ablation and retraining operate on different models, we compare the methods by checking how well visual attribution aligns with counterfactual attribution in each. Visual attribution of a generated sample is performed by first selecting a distance metric (see Appendix \ref{appdx-setup-percmetr} for the metrics we consider). Then for each data source, we compute the minimum distance between any training data arising from that data source and the generated sample. We then rank the data sources, such that the one with the smallest distance is the most attributed data source and the one with the smallest is the least.

Counterfactual attribution of a generated sample is also performed by first selecting a distance metric (for the main body of this work it is always Euclidean). Then for each counterfactual, we compute the distance between the factual and counterfactual sample. The data sources are then \emph{inversely} ranked according to the distance of their counterfactuals, such that the one with the largest distance is the most attributed data source and the one with the smallest is the least.

We check for intersections in the top-8 visually attributed (according to the Euclidean distance) and counterfactually attributed images. For ablation, out of 1024 generated samples there is an intersection 272 times. For retraining, this occurs 283 times, which is comparable. For reference, if attributions are distributed uniformly at random, then the expected number of intersections is 159 with a standard deviation of 12, so both these values are significantly above random.
Interestingly, counterfactual attribution via differential ablation is far more aligned with visual attribution, attaining a top-8 intersection 838 times. More comparisons with different visual metrics can be found in Appendix \ref{appdx-res-benchmark}, all of which show that ablation and retraining attribution are comparably aligned to visual attribution.

Visual attribution can be misleading (see Section \ref{sec-res-viscf}), so alignment with visual attribution should not be conflated with performance. Rather, our results here should be interpreted as a statement on how ABCs behave similarly to RBCs.

\subsection{Differential ablation based attribution finds visually similar images with small training sets}
\label{sec-res-diffabl}

For each of the 23 ensembles, we generated 1000 samples and attributed them to the training set using counterfactual attribution via differential ablation. The results are visually presented in Figure \ref{fig5}. We find that on ensembles trained on small training sets like MNIST\_256, CIFAR-10\_500, CelebA\_SD\_512, or MetFaces, the attributed training data bears visual similarity to the generated sample. However, at larger training set sizes the attributions become less visually intuitive.

We quantify this effect by reporting the \emph{visual similarity rank} of the attributed data source. This is calculated by taking each data source and calculating the minimum distance between training points in it and the generated sample. The data sources are then ranked, and the rank is normalized to be between 0 and 1, so that the data source with a 0 produced the image that is the most visually similar to the generated sample. Distributions of visual similarity rank are reported in Figure \ref{fig5} (for visual similarity ranks calculated with the Euclidean distance) and Appendix \ref{appdx-res-diffablres} (for other perceptual metrics). These quantitative results also show that there tends to be strong visual similarity when training sizes are small, which deteriorates as training sets increase.

\ifrenderfigures
\begin{figure}
  \begin{center}
  \includegraphics[width=0.48\linewidth]{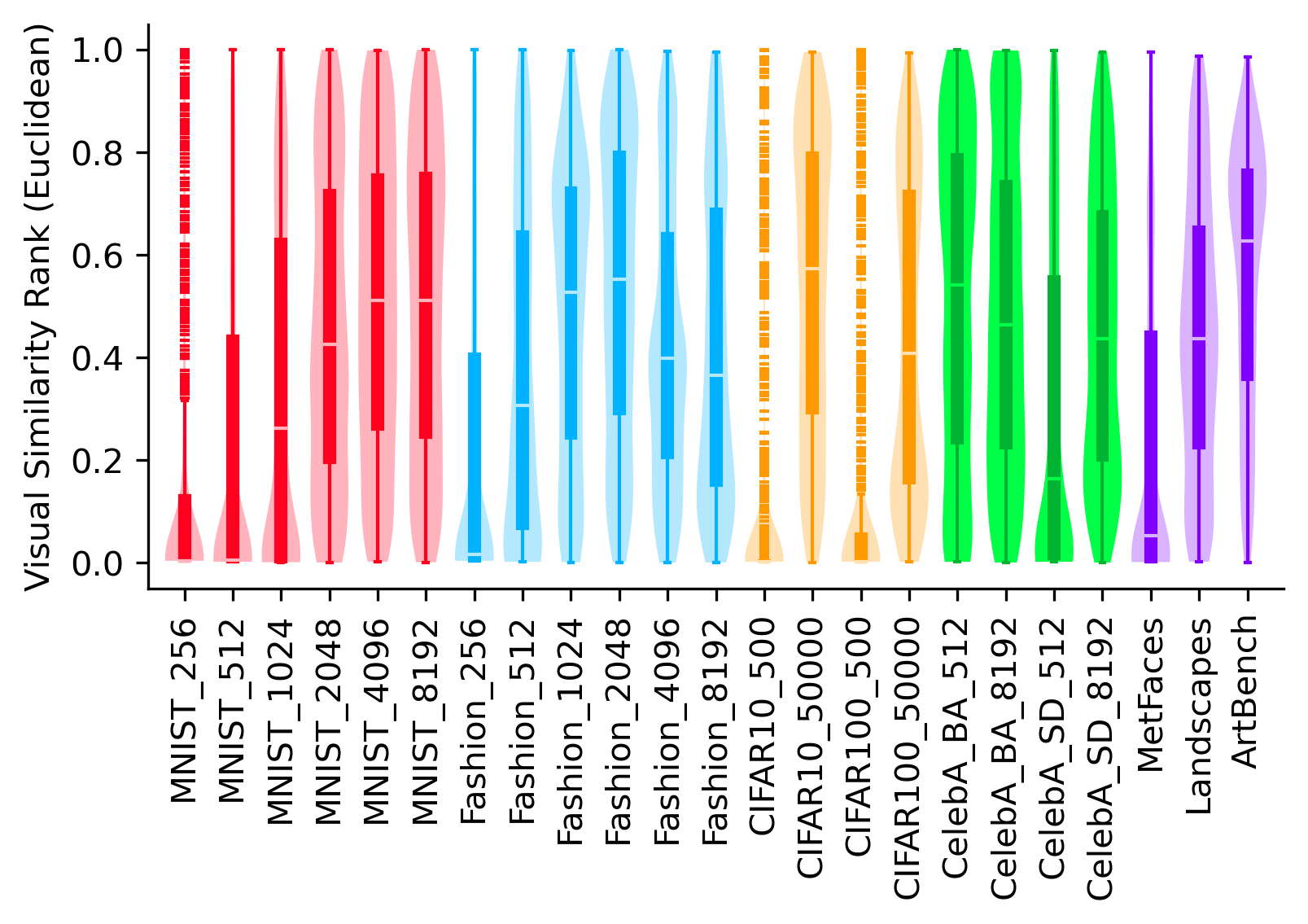}
  \includegraphics[width=0.48\linewidth]{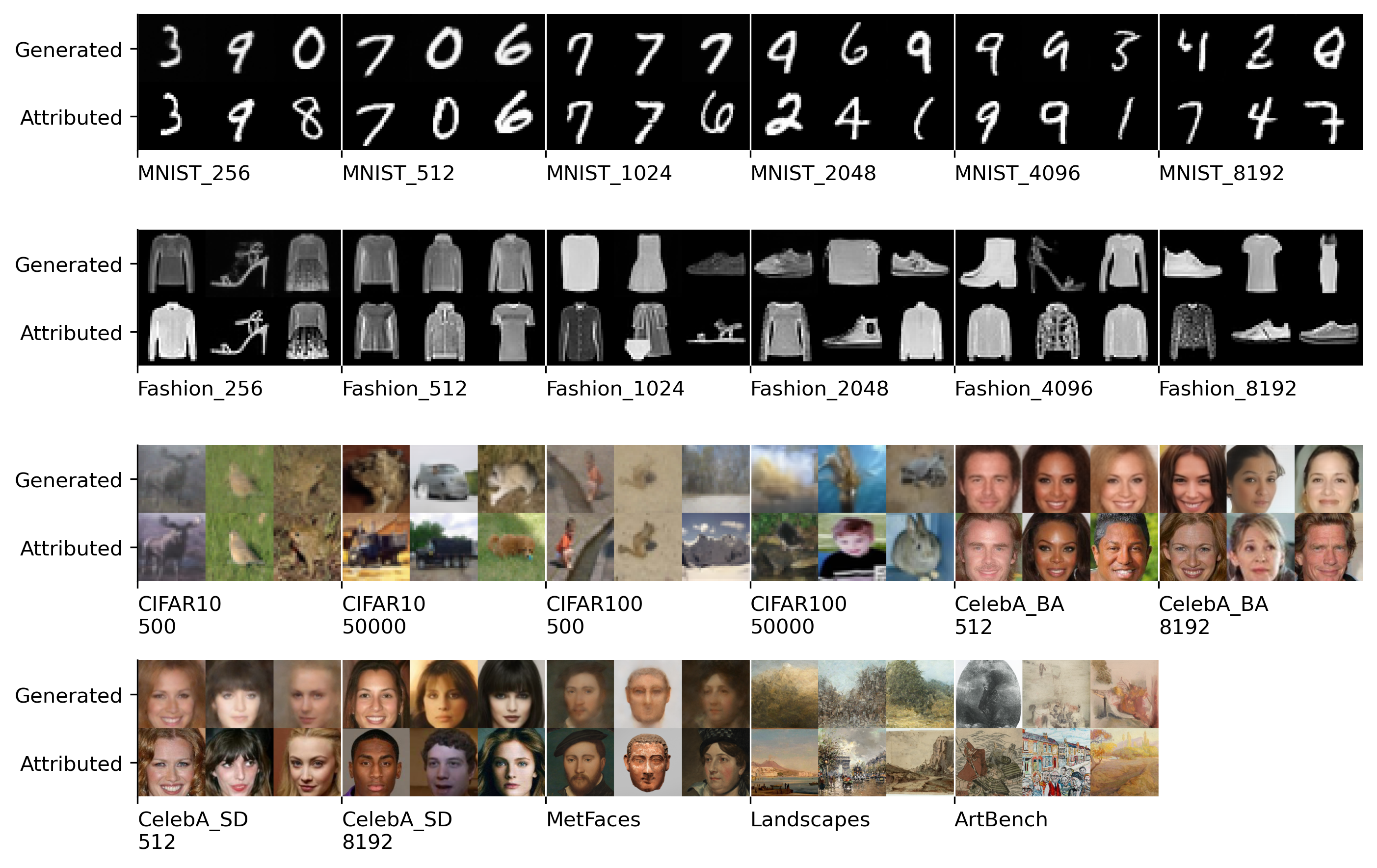}
  \end{center}
  \caption{Differential ablation attributes visually similar images when training set is small}
  On the left panel we show the distribution of visual similarity ranks of top counterfactually attributed sample using violin and box plots. On the right we visually compare select generated samples with their top counterfactually attributed data source. If a data source is responsible for multiple points, the one with the smallest Euclidean distance to the factual sample is selected for display. 
  \label{fig5}
\end{figure}
\fi

\subsection{Visual attribution and counterfactual attribution diverges at large training set sizes}
\label{sec-res-viscf}

For each of the 23 ensembles, we generated 100 samples and their full counterfactual landscapes, with the exception of MNIST\_8192 and FASHION\_8192 where we generated 50, and CIFAR-10\_50000 and CIFAR-100\_50000 where we generated 6. For each factual sample we then ranked the counterfacuals by their Euclidean distance to the factual sample and scaled the ranks so all values are between 0 and 1, with 0 being close to the factual sample and 1 being far. We refer to this score as the counterfactual's \emph{counterfactual distance rank}.

We then visually attribute the factual samples to the training set, and computed the counterfactual distance rank of the counterfactual that is associated to the data source that produced the attributed training point. The distributions are provided in Figure \ref{fig6} (for visual attributions that are made using the Euclidean distance, for other visual attributions see Appendix \ref{appdx-res-cdr}), where it can be observed that while these ranks tend to be high for smaller training sets, the distribution of these ranks become increasingly uniform as training set sizes increase. We also show in Figure \ref{fig6} that at larger training sets, even when a similar image to the generated sample is found in the training set, the counterfactual where it is missing does not necessarily change much, implying that the visually similar image did not actually play a significant causal role in the creation of the sample.

\ifrenderfigures
\begin{figure}
  \begin{center}
  \includegraphics[width=0.45\linewidth]{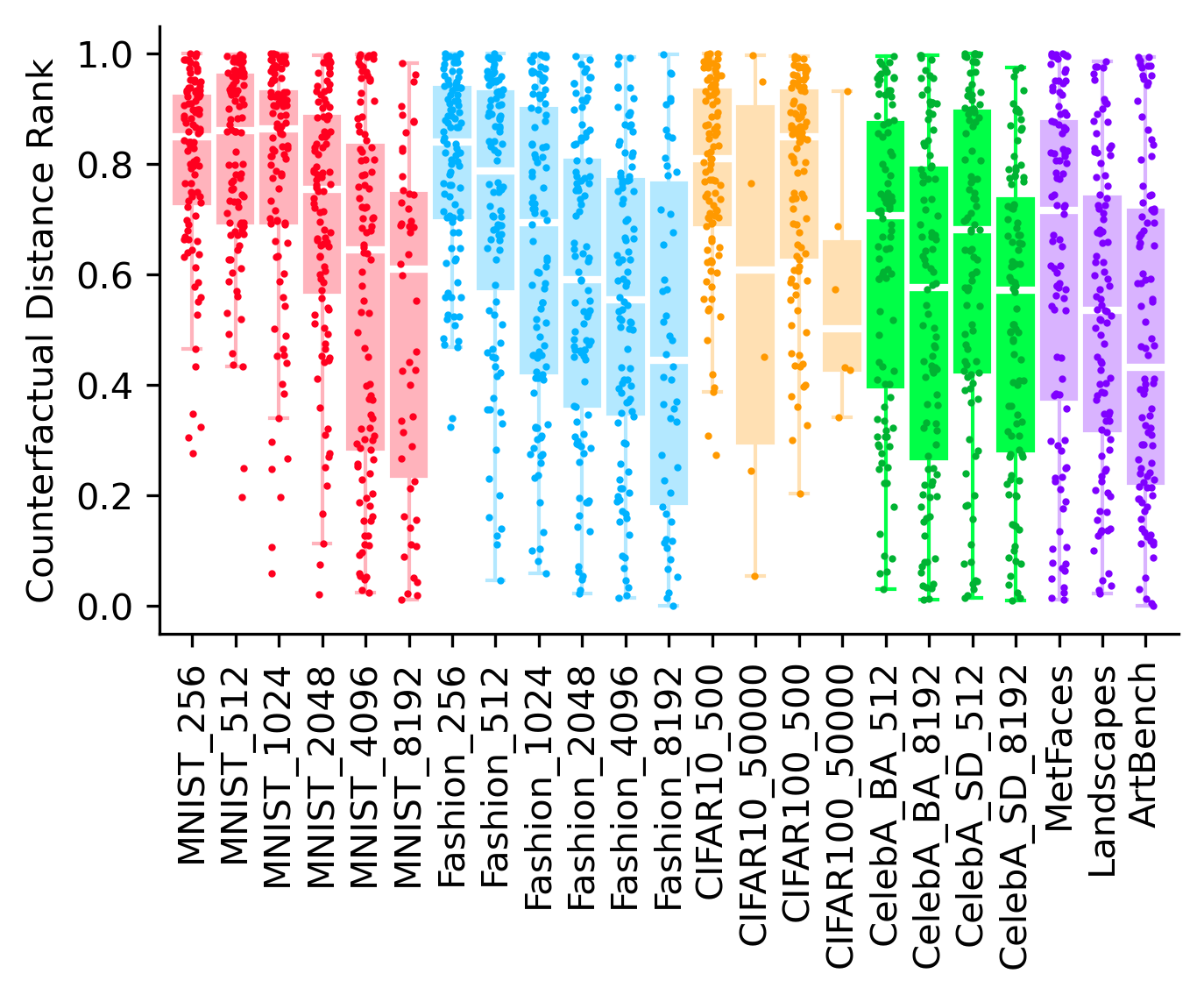}
  \includegraphics[width=0.5\linewidth]{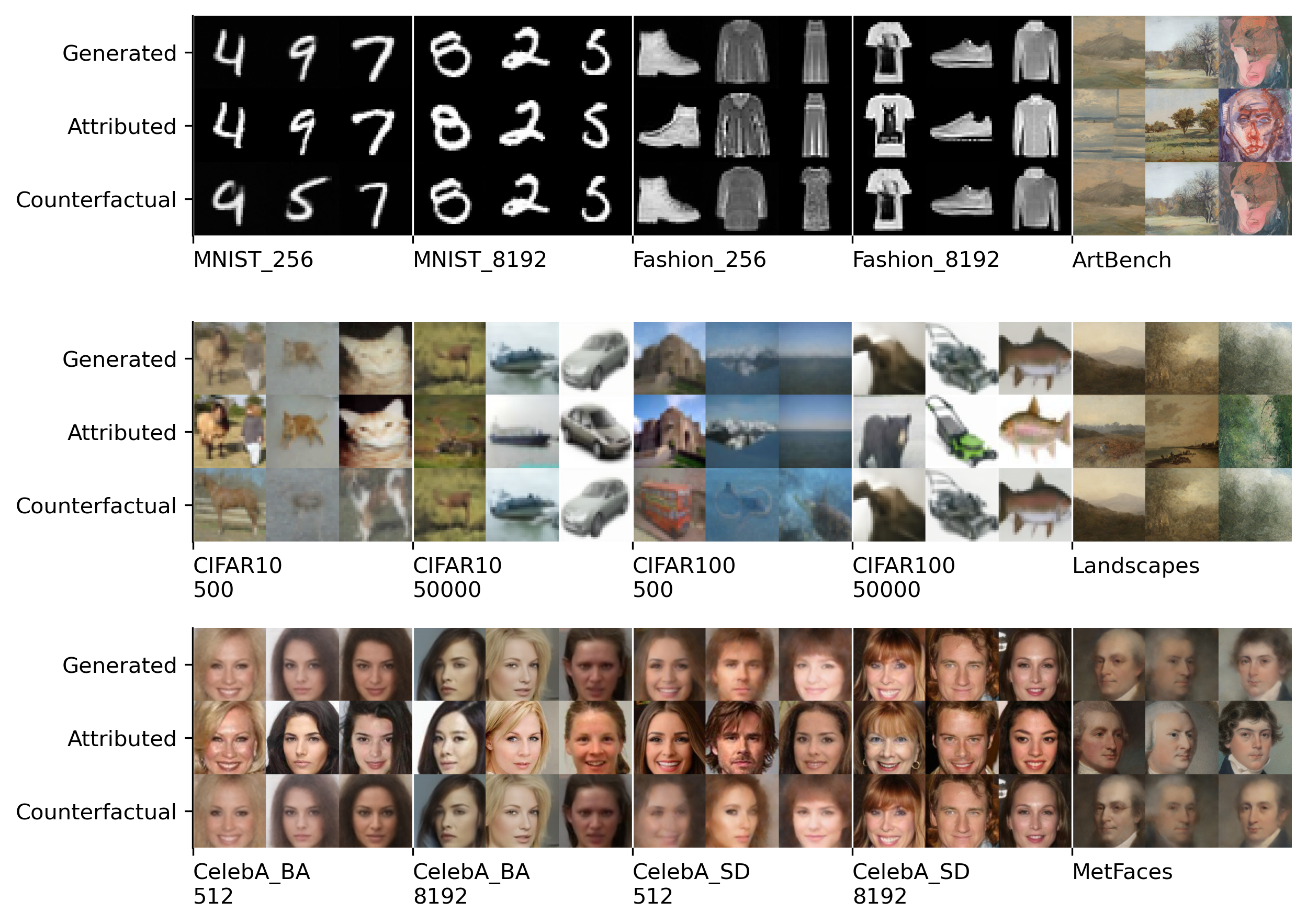}
  \end{center}
  \caption{Visually similar training images are not necessarily influential when training set is large}
  On the left panel we show the distribution of counterfactual distance ranks of top visually attributed samples using strip and box plots. On the right we compare generated samples to the top visually attributed sample and the counterfactual sample that arises if the visually attributed sample were removed. Note how in MNIST\_8192, despite finding very close images in the training set, removing them does not perceptibly change the counterfactual sample.
  \label{fig6}
\end{figure}
\fi

Great care should be taken when interpreting attribution results, since it is tempting to view visual similarity as a form of ground truth for attribution.

\subsection{Generated samples can be unattributable}

We show that as training sets increase in size, generated samples can become fundamentally unattributable. First we define the \emph{counterfactual radius}:

\begin{definition}
The counterfactual radius of a generated sample is the maximum distance between it and a member of its counterfactual landscape.
\end{definition}

The counterfactual radius captures the ``attributability'' of a generated sample. If a counterfactual lies far from the sample, then it is possible that the training data associated with that counterfactual was ultimately highly responsible for directing the sampling process. However, if all counterfactuals lie close to the sample, then it becomes difficult to claim that any part of the training data was responsible for causing the sampling process to evolve the way it did. 

Specifically, consider the case when the counterfactual radius of a sample is zero and the entire counterfactual landscape collapses into that sample. Then any claim of attribution of that sample to some data source can be refuted: each counterfactual acts as a ``certificate of non-attributability'', demonstrating that the data source in question was not needed to generate the sample. We will refer to such samples as \emph{unattributable samples}.

True unattributable samples can arise when we generate samples in a discrete space. In Figure \ref{fig8}, we show true attributable samples that are generated by an ensemble which is trained to generate MNIST digits where each pixel is either black or white. These are rare, accounting for only 14 of 3731 generated samples.

\ifrenderfigures
\begin{figure}
  \begin{center}
  \includegraphics[width=0.45\linewidth]{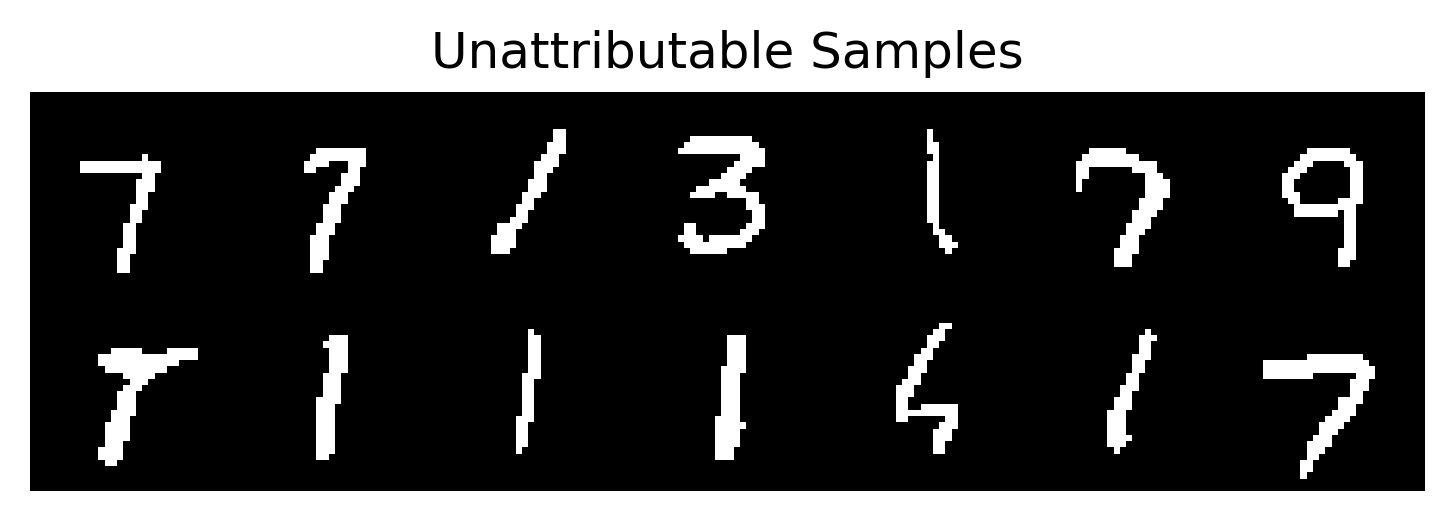}
  \includegraphics[width=0.28\linewidth]{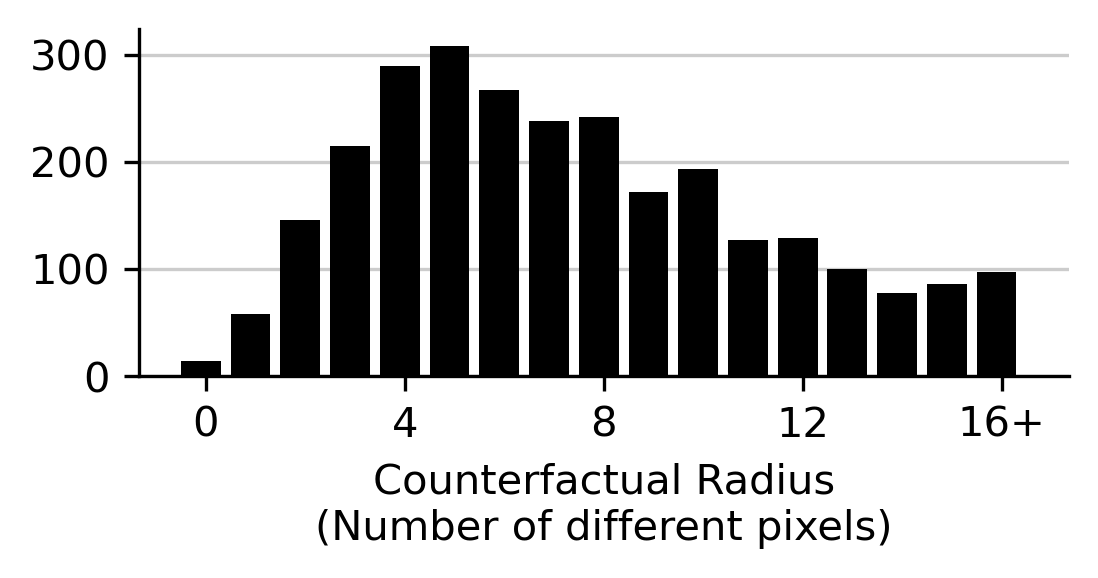}
  
  \includegraphics[width=0.48\linewidth]{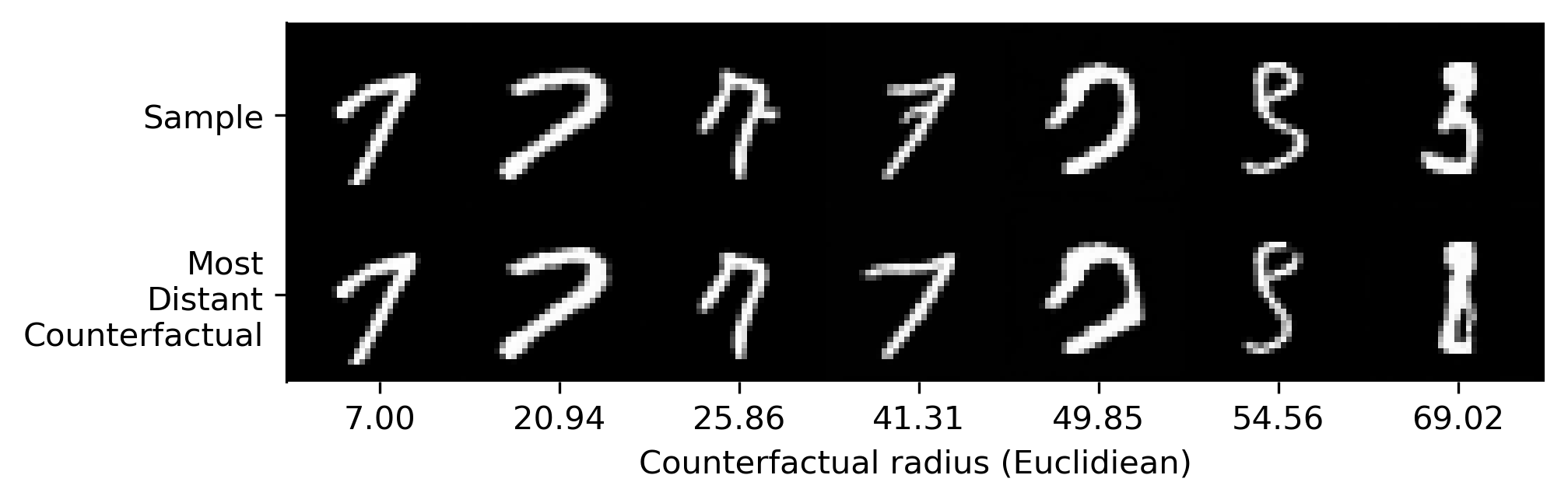}
  \includegraphics[width=0.48\linewidth]{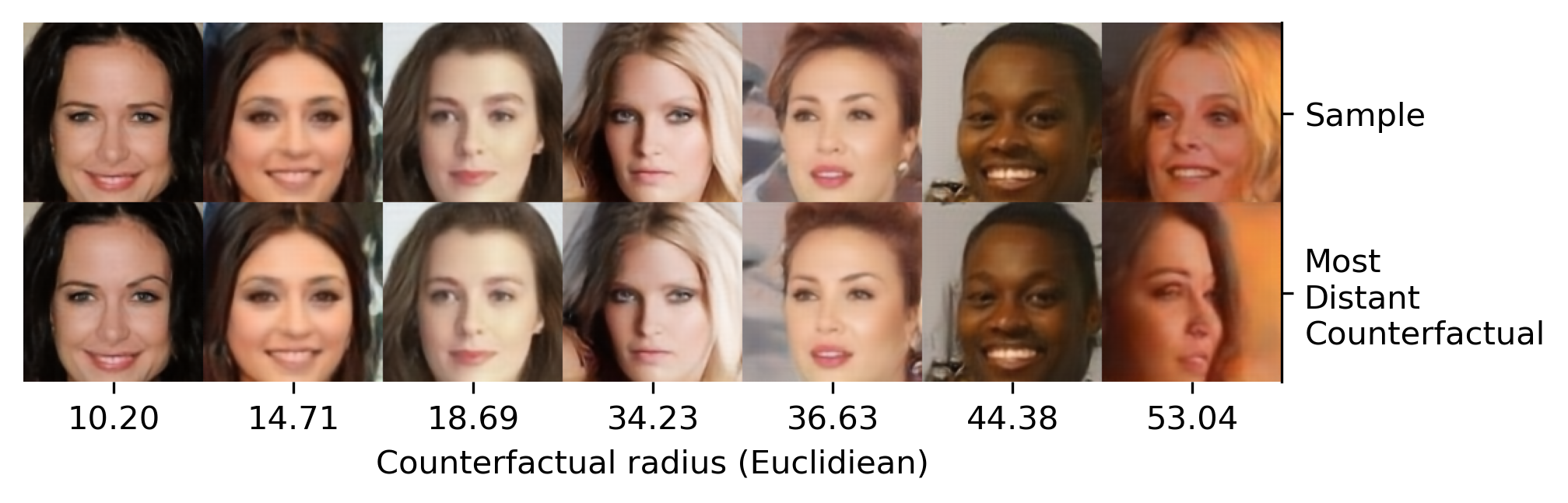}

  \end{center}
  \caption{Unattributable and nearly unattributable samples}
  The top left panel shows the 14 true unattributable samples generated from a model trained to sample binary MNIST digits. The distribution of counterfactual radii of samples generated by the model is given the in the top right panel. The two bottom panels show how distinct samples are from their most distanct counterfactuals (evaluated in Euclidean distance) at various counterfactual radii.
  \label{fig8}
\end{figure}
\fi

True attributable samples are unlikely to arise in ensembles that generate samples in a continuous image space. Therefore, it is more useful to consider some notion of ``near unattributability'', which ideally captures the idea that no element of a sample's counterfactual landscape is perceptibly different from the original sample. To operationalize this, we propose defining a sample as nearly unattributable if its counterfactual radius in under some threshold $\tau$. The metric used to measure the radius and the specific value of $\tau$ must ultimately depend on context, and for reference we present some visual comparisons between samples and their counterfactuals at varying counterfactual radii (measured in Euclidean distance) in Figure \ref{fig8}, generated by ensembles that are described in Appendix \ref{appdx-otherModels-moreens}. In addition, we present the full counterfactual landscape of an MNIST digit and CelebA face in Appendix \ref{appdx-largefig-3}, where every counterfactual differs from the factual sample by only a imperceptible amount.

\subsection{Attributability diminishes with increasing training set size}

Generating samples of low counterfactual radius is a phenomenon that occurs naturally as training sets increase in size. We present the sizes of counterfactual radii using the landscapes computed in Section \ref{sec-res-viscf} in Figure \ref{fig7} (measured in Euclidean distance, for other distances see Appendix \ref{appdx-res-cfr}). We find that there is a strong relation between the training set size and the average counterfactual radius of generated samples, which we also present in Figure \ref{fig7}. Extrapolating this relation, having $10^8$ training samples would yield an ensemble that generates samples with counterfactual radii around 7 (measured in Euclidean distance), which is very impreceptible going by the reference in Figure \ref{fig8}.

\ifrenderfigures
\begin{figure}
  \begin{center}
  \includegraphics[width=0.5\linewidth]{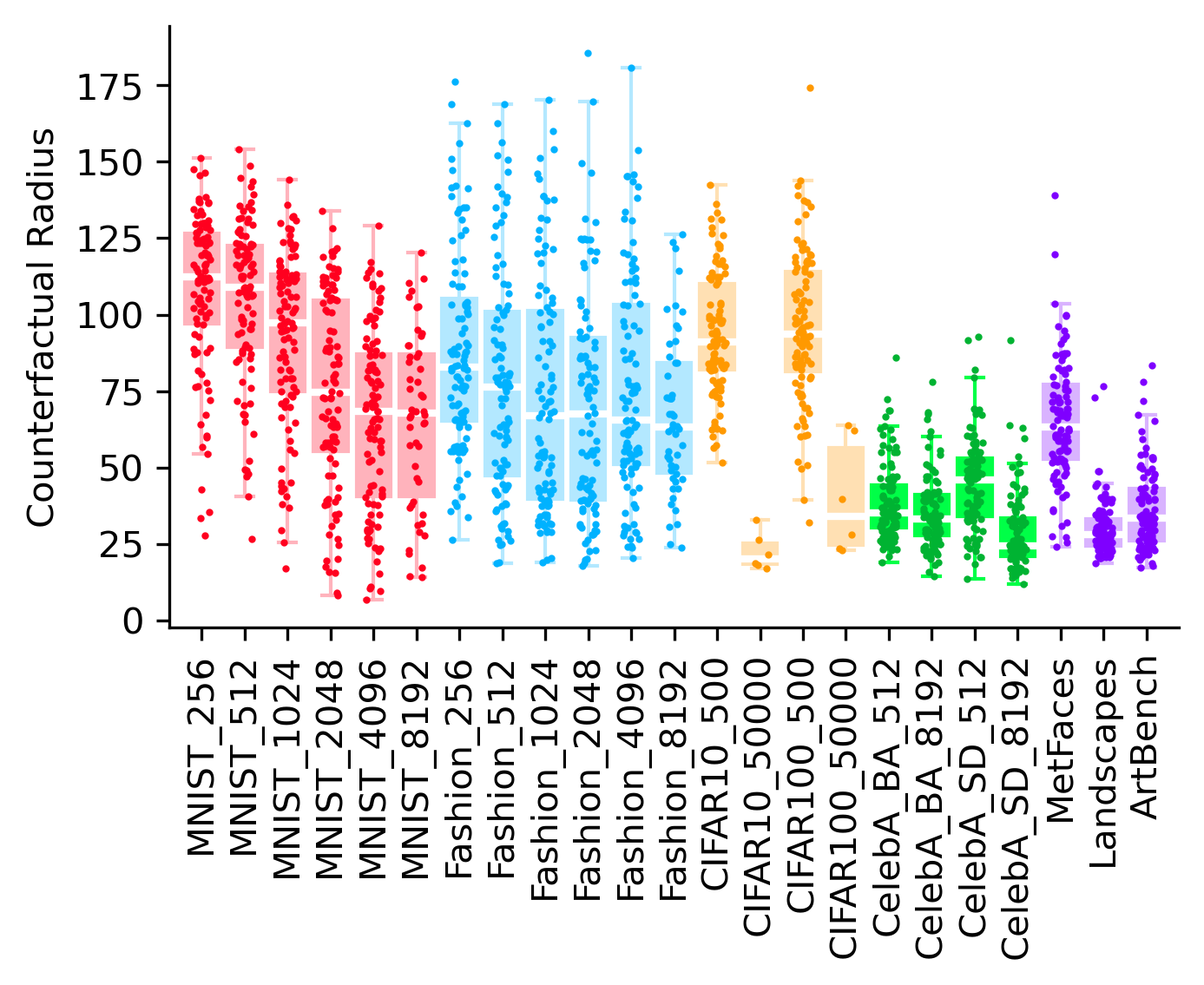}
  \includegraphics[width=0.4\linewidth]{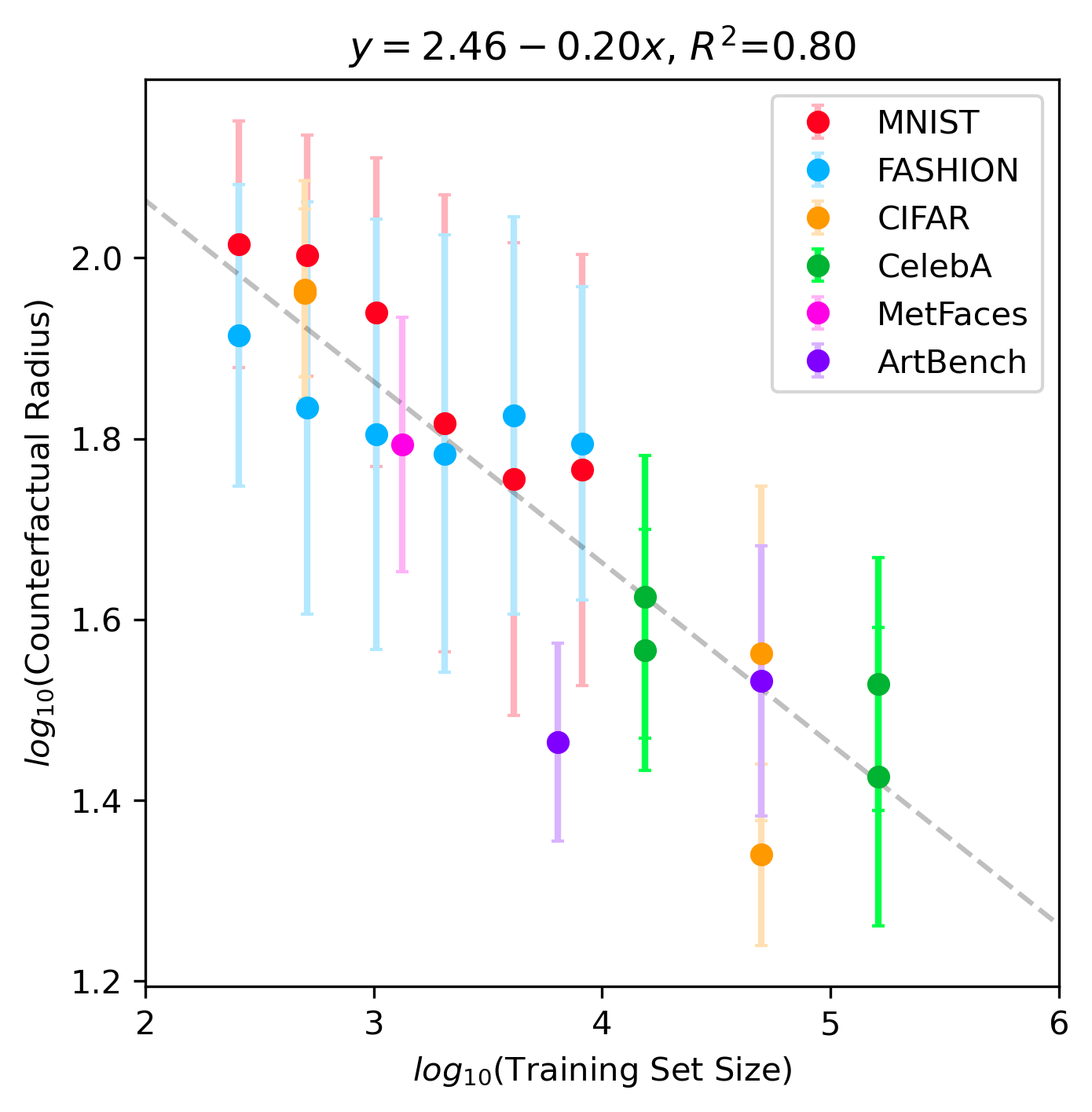}
  \end{center}
  \caption{Counterfactual radius drops with increasing training set size}
  On the left panel we show the distribution of counterfactual radii using strip and box plots. Images were scaled to 3x256x256 before computing Euclidean distance to ensure distances are comparable across datasets. On the right we plot for each ensemble the average log$_{10}$ counterfactual radii of samples generated by the ensemble against log$_{10}$ of the size of its training set. Color is used to indicate type of the training dataset, and error bars provide the standard deviation of the distribution of log$_{10}$ counterfactual radii. The line of best fit is reported along with $R^2$, which is statistically significant ($p\leq 8.76*10^{-9}$).
  \label{fig7}
\end{figure}
\fi

This suggests that the divergence between counterfactual and visual attribution observed at large training set sizes (Section \ref{sec-res-viscf}), and the visually unintuitive attributions made through differential ablation (Section \ref{sec-res-diffabl}) are due to a foundational collapse in counterfactual attributability.

The existence of unattributable and nearly unattributable samples has serious scientific and policy consequences, especially given that they occur naturally as a consequence of larger training set sizes. Scientifically, unattributability dismantles the foundation of leave-one-out counterfactual analysis for models trained on large datasets, necessitating the pursuit of alternative paradigms. Policywise, it is possible that unattributable samples circumvent existing copyright standards, since an unattributable sample needs no \emph{access} to any given piece of training data to be generated.
\section{Conclusion}
We present ablation as an alternative to retraining as a paradigm for computing counterfactuals. The main advantage of this approach is that no retraining is required, which allows us to for the first time compute entire counterfactual landscapes exactly. We analyze these landscapes and discover that the foundations underlying leave-one-out counterfactual analysis of assessing influence diminishes at large training set sizes.

\section*{Additional resources}
Additional resources like scripts for this paper can be accessed from our project page at \url{https://zheng-dai.github.io/AblationBasedCounterfactuals/}. Publically accessible datasets used for experiments are listed in Table \ref{table-datasets}.

\newpage
\appendix
\section*{Appendix}
\section{Theoretical analysis}
\subsection{Proof of Theorem \ref{thm-1}}
\label{appdx-proofs-thm1}
Suppose by way of contradiction that $S(s) \setminus S(s')$ is the empty set. Then $S(s') \subseteq S(s)$. But because each codeword has equal Hamming weight, it must be the case that $|S(s')| = |S(s)|$, and since we are working with finite sets that means $S(s') = S(s)$, which can only happen if $s' = s$ because each data source is assigned a unique codeword. This contradicts the premise that $s$ and $s'$ are distinct, which proves the statement.

\subsection{Given $N$ data sources, it suffices to train $\mathcal{O}(log(N))$ models}
\label{appdx-proofs-sperner}

There exist $\binom{n}{\lfloor n/2 \rfloor}$ unique binary vectors of length $n$ with Hamming weight $\lfloor n/2 \rfloor$. Therefore, if $N \leq \binom{n}{\lfloor n/2 \rfloor}$, an ensemble of size $n$ suffices to account for all $N$ data sources. Since $\binom{n}{\lfloor n/2 \rfloor}$ is $\mathcal{O}(2^n/\sqrt{n})$, it suffices for $n$ to be $\mathcal{O}(log(N))$.

\subsection{Runtime analysis}
\label{appdx-proofs-runtime}

To fully capture all the amortizations that can be leveraged, we consider the task of generating $n$ samples and their full counterfactual landscapes when analyzing runtime.

Let $T$ be the runtime for training a model. Let $t$ be the runtime of sampling the diffusion model. Let $N$ be the number of data sources.

In the retraining paradigm, we would first need to train $N+1$ models, one for the factual scenario where all the training data is present and $N$ for each counterfactual scenario where a data source is missing. Training all of these models would take time $\mathcal{O}(NT)$. Once all of these models are trained, for each of the $n$ samples each of these models would need to be individually run, taking time $\mathcal{O}(Nt)$. Altogether, to the evaluate the landscapes of $n$ samples would take time $\mathcal{O}(nNt)$. Altogether, this task would take time \hl{$\mathcal{O}(NT + nNt)$}.

In the ablation based paradigm, we would first need to train $\mathcal{O}(log(N))$ models to create the ensemble, which would take time $\mathcal{O}(log(N)T)$. Once trained, for each of the $n$ samples we would need to ablate and run the ensemble. Running the ensemble takes time $\mathcal{O}(log(N)t)$, so running it $N+1$ times for the factual and $N$ counterfactual scenarios takes time $\mathcal{O}(Nlog(N)t)$. Altogether, to the evaluate the landscapes of $n$ samples would take time $\mathcal{O}(nNlog(N)t)$. Altogether, this task would take time \hl{$\mathcal{O}(log(N)T + nNlog(N)t)$}.

Due to the $\mathcal{O}\log(n)$ overhead introduced by the ensemble in the ablation based paradigm, it is theoretically possible that in the limit of very large $n$ the ablation paradigm has a worse runtime. However, the $\mathcal{O}(NT)$ present in the runtime of the retraining paradigm represents training a new model for each data source, which would represent years or even centuries worth of compute (training a model could take days to weeks, and the number of data sources could range up to $10^3$ or $10^6$ or even more). Therefore, for most reasonable scenarios, the ablation based paradigm represents a large improvement, since the $\mathcal{O}(NT)$ training time is reduced to $\mathcal{O}(log(N)T)$.

We also analyze the runtime of approximating this landscape with differential ablation. Let $t'$ be size of the sample (i.e. number of dimensions). We require time $\mathcal{O}(log(N)T)$ to compute the ensemble. For each sample, we require time $\mathcal{O}(log(N)t)$ to generate a column of the Jacobian using a round of forward mode automatic differentiation. The Jacobian has $\mathcal{O}(log(N))$ columns, so computing the Jacobian takes time  $\mathcal{O}(log^2(N)t)$. Once the Jacobian is computed, each counterfactual is approximated with a vector matrix product, which takes time propertional to the size of the Jacobian, which is $\mathcal{O}(log(N)t')$. Thus evaluating this for the entire landscape takes time $\mathcal{O}(Nlog(N)t')$. Altogether, the task takes time \hl{$\mathcal{O}(log(N)T+nlog^2(N)t + nNlog(N)t')$}.

The runtime of $log(N)t'$ arises from computing a matrix-vector product. This is an extremely optimized operation, and is generally much faster than the runtime of $t$ that is needed for sampling. If we set this operation to $\mathcal{O}(1)$, we find that the runtime reduces to \hl{$\mathcal{O}(log(N)T+n(log^2(N)t + N))$}, which is more representative of the true runtime of differential ablation.
\section{Experimental setup}
\label{appdx-experimental-setup}
\subsection{Data preparation}
\label{appdx-exp-data-prep}
We used various datasets in our experiments, which are summarized in Table \ref{table-datasets}.

\begin{table}[h]
\label{table-datasets}
\caption{Summary of datasets that were used for our experiments}
\label{table0}
\vskip 0.15in
\begin{center}
\begin{small}
\begin{tabular}{l c c c c}
\toprule
Dataset name & License & Reference & Image size & Source \\
\midrule
MNIST & MIT & \citet{lecun2010mnist} & 1x28x28 & TorchVision\\
FASHION & MIT & \citet{xiao2017fashion} & 1x28x28 & TorchVision\\
CIFAR-10 & unknown & \citet{krizhevsky2009learning} & 3x32x32 & TorchVision\\
CIFAR-100 & unknown & \citet{krizhevsky2009learning} & 3x32x32 & TorchVision\\
CelebA & \href{https://mmlab.ie.cuhk.edu.hk/projects/CelebA.html}{custom} & \citet{liu2018large} & various & TorchVision\\
MetFaces & CC BY-NC 2.0 & \citet{karras2020training} & 3x1024x1024 & \href{https://github.com/NVlabs/metfaces-dataset}{GitHub}\\
ArtBench & MIT & \citet{liao2022artbench} & 3x256x256 & \href{https://github.com/liaopeiyuan/artbench/tree/main}{GitHub}\\
\bottomrule
\end{tabular}
\end{small}
\end{center}
\vskip -0.1in
\end{table}

These datasets are processed in the following way: MNIST and Fashion-MNIST (henceforth referred to as FASHION for brevity) was padded to be 32x32 by adding a two-pixel wide black border. CelebA was center cropped to be 128x128. MetFaces was scaled to 256x256 via bilinear interpolation.

Subsets of these models were used to train various ensembles. For MNIST and FASHION, we selected subsets of size 256, 512, 1024, 2048, 4096, and 8192 from their test splits. For CIFAR-10 and CIFAR-100, we selected subsets of size 500 and 50000 from their training splits. For CelebA, we selected subsets of size 15363, 15365, and 162770 from its training splits. For MetFaces, we used the entire dataset, and for ArtBench, we took subsets of size 6414 and 50000 from its training set. The subset of 6414 was selected to be landscape pictures, and was selected by computing the OpenCLIP embedding and computing the cosine similarity between it and the OpenCLIP embeddings of the phrases ``a realistic high quality landscape painting'' and ``a blurry mess''. Cosine similarities of 0.25 or larger with the first phrase and strictly less than 0.2 for the second phrase were selected for inclusion, and this subset is referred to as ``Landscapes''. The ``ViT-H-14-378-quickgelu'' model pretrained on ``dfn5b'' was used for computing OpenCLIP embeddings.

Data was prepared for latent diffusion for images in CelebA, MetFaces, and ArtBench, while pixel space diffusers were trained for the other datasets. Most datasets were emedded with the Stable-Diffusion-v1-5 autoencoder~\cite{Rombach_2022_CVPR}. In the case of CelebA, images were upscaled to 256x256 with bilinear interpolation before embedding.

For CelebA, another basic autoencoder was trained in-house (see Appendix \ref{appdx-otherModels-ba}) on the test split of the CelebA dataset. The autoencoder maps between images of shape 3x128x128 and a latent space of shape 4x32x32. A second set of embeddings was produced using this autoencoder.

For ArtBench and MetFaces, embeddings of flipped versions of each image was also computed to augment the dataset. The flipped image is considered to arise from the same data source as the original image.

\subsection{Training ensembles}
\label{appdx-models-training}

A total of 23 ensembles were trained. A summary of the data used to train each model is provided in Table \ref{table-models}.

\begin{table}[h]
\caption{Summary of the ensembles that were trained}
\label{table-models}
\vskip 0.15in
\begin{center}
\begin{small}
\begin{tabular}{l C{1.5cm} C{1cm} c c C{2cm}}
\toprule
Ensemble name & Training set origin & Training set size & Latent space & Data Source & Counterfactual landscape size \\
\midrule
MNIST\_256 & MNIST & 256& Pixel & Data Point & 256 \\
MNIST\_512 & MNIST & 512 & Pixel & Data Point & 512 \\
MNIST\_1024 & MNIST& 1024 & Pixel & Data Point & 1024  \\
MNIST\_2048 & MNIST& 2048 & Pixel & Data Point & 2048  \\
MNIST\_4096 & MNIST & 4096 & Pixel & Data Point & 4096 \\
MNIST\_8192 & MNIST & 8192 & Pixel & Data Point & 8192 \\
FASHION\_256 & FASHION & 256 & Pixel & Data Point & 256 \\
FASHION\_512 & FASHION & 512 & Pixel & Data Point & 512 \\
FASHION\_1024 & FASHION & 1024 & Pixel & Data Point & 1024 \\
FASHION\_2048 & FASHION & 2048 & Pixel & Data Point & 2048 \\
FASHION\_4096 & FASHION & 4096 & Pixel & Data Point & 4096 \\
FASHION\_8192 & FASHION & 8192 & Pixel & Data Point & 8192 \\
CIFAR-10\_500 & CIFAR-10 & 500 & Pixel & Data Point & 500 \\
CIFAR-10\_50000 & CIFAR-10 & 50000 & Pixel & Data Point & 50000 \\
CIFAR-100\_500 & CIFAR-100 & 500 & Pixel & Data Point & 500 \\
CIFAR-100\_50000 & CIFAR-100 & 50000 & Pixel & Data Point & 50000 \\
CelebA\_BA\_512 & CelebA & 15363 & Basic autoencoder & Celebrity & 512 \\
CelebA\_BA\_8192 & CelebA & 162770 & Basic autoencoder & Celebrity & 8192 \\
CelebA\_SD\_512 & CelebA & 15365 & Stable Diffusion & Celebrity & 512 \\
CelebA\_SD\_8192 & CelebA & 162770 & Stable Diffusion & Celebrity & 8192  \\
MetFaces & MetFaces & 1336 & Stable Diffusion & Artist & 744 \\
Landscapes & ArtBench & 6414 & Stable Diffusion & Artist & 946 \\
ArtBench & ArtBench & 50000 & Stable Diffusion & Artist & 2108 \\
\bottomrule
\end{tabular}
\end{small}
\end{center}
\vskip -0.1in
\end{table}

Models were trained as described in \citet{ho2020denoising}. Training was done with 1000 denoising steps, while sampling was done with 50 denoising steps. Random horizontal flipping was used as a data augmentation strategy when training on data from FASHION, CIFAR-10, CIFAR-100, MetFaces, and ArtBench. Adam was used for gradient descent with default PyTorch~\cite{paszke2019pytorch} parameters, a learning rate of 0.0001, and a batch size of 128. Two variants of the U-Net was used based on the \href{https://huggingface.co/docs/diffusers/v0.27.2/en/api/models/unet2d}{Diffusers} implementation. Each U-net consisted of 4 down and upblocks, with an attention layer on the second down and third upblock. The smaller U-Net had channel sizes (128, 256, 512, 512), while the larger U-Net had channel sizes (192, 384, 768, 768). The smaller U-Net was trained on MNIST, FASHION, CelebA, MetFaces, and Landscapes, while the larger U-Net was trained on CIFAR-10, CIFAR-100, and ArtBench. A dropout of 0.1 was applied to models trained on CIFAR-10, CIFAR-100, and ArtBench (but not on the Landscapes subset).

Each training dataset was individually normalized so that the mean of an arbitrary value in the dataset tensor is 0 and the standard deviation is 1. The exception is MNIST and FASHION, where values were transformed by subtracting 0.5 and multiplying by 2, and CIFAR-10 and CIFAR-100, where values were transformed by subtracting 1 and multiplying by 4, and CelebA embeddings embedded by the basic autoencoder, where each channel was normalized individually, instead of the entire dataset tensor.

Each ensemble was trained to contain 22 members, with the exception of Landscapes and MetFaces, which contained 24 members. Codes were assigned to data sources (see Section \ref{sec-meth-encoded}) such that each model in the ensemble sees approximately half the training set.

\subsection{Additional ensembles}
\label{appdx-otherModels-moreens}

In addition, 7 additional ensembles were trained. 2 ensembles of size 6 were trained on the entire MNIST and FASHION test splits, where each class was considered a data source. 2 ensembles of size 16 were trained on subsets of size 384 of MNIST and FASHION test splits, where each image was its own data source. These models were used to benchmark the ablation paradigm against the retraining paradigm.

For the next 2 ensembles, one was trained on the training split of MNIST and one on the latent embeddings of CelebA induced by the basic autoencoder. These ensembles contained 14 members each, and data sources were combined so the counterfactual landscape only contained 96 elements. This allows for quicker calculation of counterfactual landscapes, though at the cost of granularity. These models were used to produce examples of different counterfactual radii that were presented in Figure \ref{fig8}.

Finally, 1 ensemble with 20 members was trained on binarized MNIST digits from the training set. The binarized MNIST digits contained duplicates, which were removed. The resulting dataset contained 59984 elements, which were randomly assorted into 190 data sources, again for easier counterfactual computation. This ensemble was used to find true unattributable samples, which are presented in Figure \ref{fig8}. For ensemble and this ensemble only, the codes are assigned such that each model only sees approximately 1/10th of the dataset as opposed to the usual half.

\subsection{Basic Autoencoder}
\label{appdx-otherModels-ba}

We trained a basic autoencoder that maps between images with shape 3x128x128 and latent embeddings of shape 4x32x32. The Stable Diffusion autoencoder has been trained on a vast amount of data, which may intersect the training sets of our diffusion models. We train this autoencoder on the test split of the CelebA dataset, which ensures no overlap between images used to train the autoencoder and the diffusion model.

The encoder is made of 3 convolutions (channel widths going from 3 to 64 to 64 to 4), downscaling by a factor of 2 for the first two steps with kernel sizes of 7x7, 7x7, and 5x5. The decoder is made of 2 transposed convolutions that upscale by a factor of 2 followed by 2 convolutions. Kernel sizes are 7x7, 7x7, 5x5, and 3x3. A ReLU follows each convolution, and after the second ReLU in the encoder a BatchNorm2d is used.

This autoencoder was trained on the test split of the CelebA dataset to minimize mean squared error between the original image and the image after it has been through the autoencoder for 10 epochs with a batchsize of 64. Adam was used for gradient descent with default paramters.

After this, the loss was adjusted to incoporate patches to improve local detail. To calculate this loss, the image was divided into 8x8, 4x4, 2x2, and 1x1 patches. The mean squared loss was calculated for each patch for each patch size, and the maximum was taken over all patches. The loss was then scaled by 4, 8, 16, and 32 respectively, and added to the default mean squared error loss. Training continued for an additional 574 epochs with this adjusted loss.

\subsection{MNIST classifier}
\label{appdx-otherModels-mnist}
For classifying MNIST and FASHION images, we trained a ResNet-18~\cite{he2016deep} on the MNIST and Fashion-MNIST training split. The test split was used for validation.

Adam~\cite{kingma2014adam} was used for gradient descent using default PyTorch parameters. The models were trained normally for 50 epochs. The optimizer was then reset and the model from the epoch with the best accuracy was selected.

The selected model was trained adversarially for an additional 50 epochs using the PGDL2 adversary implemented in TorchAttacks~\cite{kim2020torchattacks}. For the adversarial phase the model was trained on perturbed and unperturbed inputs simultaneously. The attacker was permitted 50 steps with a step size of 0.2, and a raidus of 3 for MNIST and 1 for FASHION. The model from the epoch with the best adversarial accuracy was selected.

The MNIST classifier achieved an accuracy of 0.9788 and an adversarial accuracy of 0.9370 on the test split. The FASHION classifier achieved an accuracy of 0.6378 and an adversarial accuracy of 0.8077 on the test split.

\subsection{Computing visual similarity}
\label{appdx-setup-percmetr}
The Euclidean distance is often used as a surrogate for visual similarity, but it can be misleading. We therefore make use of 4 other perceptual metrics to ensure our findings are robust: LPIPS~\cite{zhang2018unreasonable}, OpenCLIP~\cite{ilharco_gabriel_2021_5143773}, and two versions of DINOv2~\cite{oquab2023dinov2}, which we term DINOv2 and DINO Patch, giving us a total of 5 vision metrics to work with.

LPIPS was run in VGG mode. The ``ViT-H-14-378-quickgelu'' model pretrained on ``dfn5b'' was used for computing OpenCLIP. OpenCLIP distances are computed by taking the cosine distance between embedded images. The ``ViT-L/14'' model with registers was used for computing DINOv2. For DINOv2, we computed the cosine distance between the class token computed by the model. For DINO Patch, we took the 256 patch tokens generated by the model, computed the Euclidean distances between corresponding tokens of two images, and took the maximum value over them.

\subsection{Computational resources}
Our experiments were run on 7 Titan RTX GPUs, each with 24190MiB of memory and 8 GeForce GTX 1080 Ti GPUs, each with 11264MiB of memory. Individual model training times varied between a few hours for MNIST models and a few days for ArtBench models. Generating a counterfactual landscape varied from a few minutes for small counterfactual landscapes of size 256 to a few days for large counterfactual landscapes of size 50000. Computing Jacobians for differential ablation were done in batches of 12-16, and takes around 2 hours.
\section{Supplemental results}

\subsection{Frechet Inception Distances of ensembles}
\label{appdx-res-fid}

For each ensemble that we trained, we computed its Frechet Inception Distance (FID)~\cite{heusel2017gans} based on 10240 samples generated from each ensemble. A lower score indicates better performance. To calculate FID, a reference dataset is required. For ensembles trained on MNIST and Fashion-MNIST, the training split of MNIST was used as reference. For ensembles trained on CIFAR-10 and CIFAR-100, the teset split of CIFAR-10 and CIFAR-100 was used as reference. For models trained on CelebA, the test split of CelebA was used as reference. For the ensemble trained on MetFaces, the MetFaces dataset of used as reference. For the ensemble trained on ArtBench, the Artbench test split was used as reference. For the ensemble trained on the Landscape subset of ArtBench, the Landscape subset was used as reference. The use of these references ensures no intersection between the data used to train the ensemble and the reference dataset for FID calculation, with the exception of MetFaces and Landscapes.

In addition to computing FIDs for the ensembles, for each ensemble we train a single diffusion model on the same training set. We compute the FID of this model as the ``single model control'' to check whether ensembling harms or improves performance. Furthermore, we also take a member of the ensemble and use it as a single diffusion model. We calculate the FID of this as the ``ensemble member control''. The summary of the results are provided Table \ref{table-fid}.

\begin{table}[h]
\caption{FIDs attained by ensembles and single models}
\label{table-fid}
\vskip 0.15in
\begin{center}
\begin{small}
\begin{tabular}{p{2.5cm} C{2cm} C{2cm} C{2cm} C{2cm}}
\toprule
& Ensemble FID & Single model FID & Ensemble member FID & Training set size \\
\midrule
MNIST\_256 & 6.245 & 4.512 & 5.474 & 256 \\
MNIST\_512 & 5.909 & 3.882 & 4.448 & 512 \\
MNIST\_1024 & 5.696 & 3.924 & 3.422 & 1024 \\
MNIST\_2048 & 4.542 & 3.340 & 3.114 & 2048 \\
MNIST\_4096 & 2.134 & 3.206 & 2.426 & 4096 \\
MNIST\_8192 & 1.995 & 3.675 & 2.466 & 8192 \\
FASHION\_256 & 7.236 & 6.815 & 7.352 & 256 \\
FASHION\_512 & 6.735 & 6.070 & 5.126 & 512 \\
FASHION\_1024 & 6.160 & 4.451 & 4.837 & 1024 \\
FASHION\_2048 & 4.816 & 4.541 & 4.381 & 2048 \\
FASHION\_4096 & 3.733 & 5.013 & 5.247 & 4096 \\
FASHION\_8192 & 4.200 & 6.523 & 5.045 & 8192 \\
CIFAR-10\_500 & 8.810 & 7.244 & 9.297 & 500 \\
CIFAR-10\_50000 & 5.017 & 4.163 & 5.166 & 50000 \\
CIFAR-100\_500 & 9.393 & 7.656 & 10.062 & 500 \\
CIFAR-100\_50000 & 5.735 & 4.418 & 5.984 & 50000 \\
CelebA\_BA\_512 & 8.214 & 5.997 & 6.085 & 15363 \\
CelebA\_BA\_8192 & 4.134 & 4.676 & 3.904 & 162770 \\
CelebA\_SD\_512 & 7.737 & 4.890 & 4.869 & 15365 \\
CelebA\_SD\_8192 & 3.380 & 3.877 & 3.654 & 162770 \\
MetFaces & 9.948 & 7.327 & 4.309 & 1336 \\
Landscapes & 11.676 & 7.858 & 8.896 & 6414 \\
ArtBench & 6.715 & 6.463 & 7.062 & 50000 \\
\bottomrule
\end{tabular}
\end{small}
\end{center}
\vskip -0.1in
\end{table}

FID calculation was performed by scaling all the images to 3x299x299 with bilinear interpolation. The channels are then normalized to by subtracting mean [0.485, 0.456, 0.406] and dividing by standard deviation [0.229, 0.224, 0.225]. The final layer was computed with the Inception\_V3 implementation in PyTorch with \href{https://pytorch.org/vision/main/models/generated/torchvision.models.inception_v3.html#torchvision.models.Inception_V3_Weights}{``IMAGENET1K\_V1''} weights.

\subsection{Differential ablation accurately predicts the effects of ablation}
\label{appdx-res-diffabl}

For each ensemble we generated 100 samples and their full counterfactual landscapes, with the exception of MNIST\_8192 and FASHION\_8192 where we generated 50, and CIFAR-10\_50000 and CIFAR-100\_50000 where we generated 6. We then ran differential ablation on each of the original samples. We find that differential ablation is able to both capture features of the overall counterfactual landscape and of individual counterfactuals (see Figure \ref{fig4}).

\ifrenderfigures
\begin{figure}
  \begin{center}
  \includegraphics[width=0.8\linewidth]{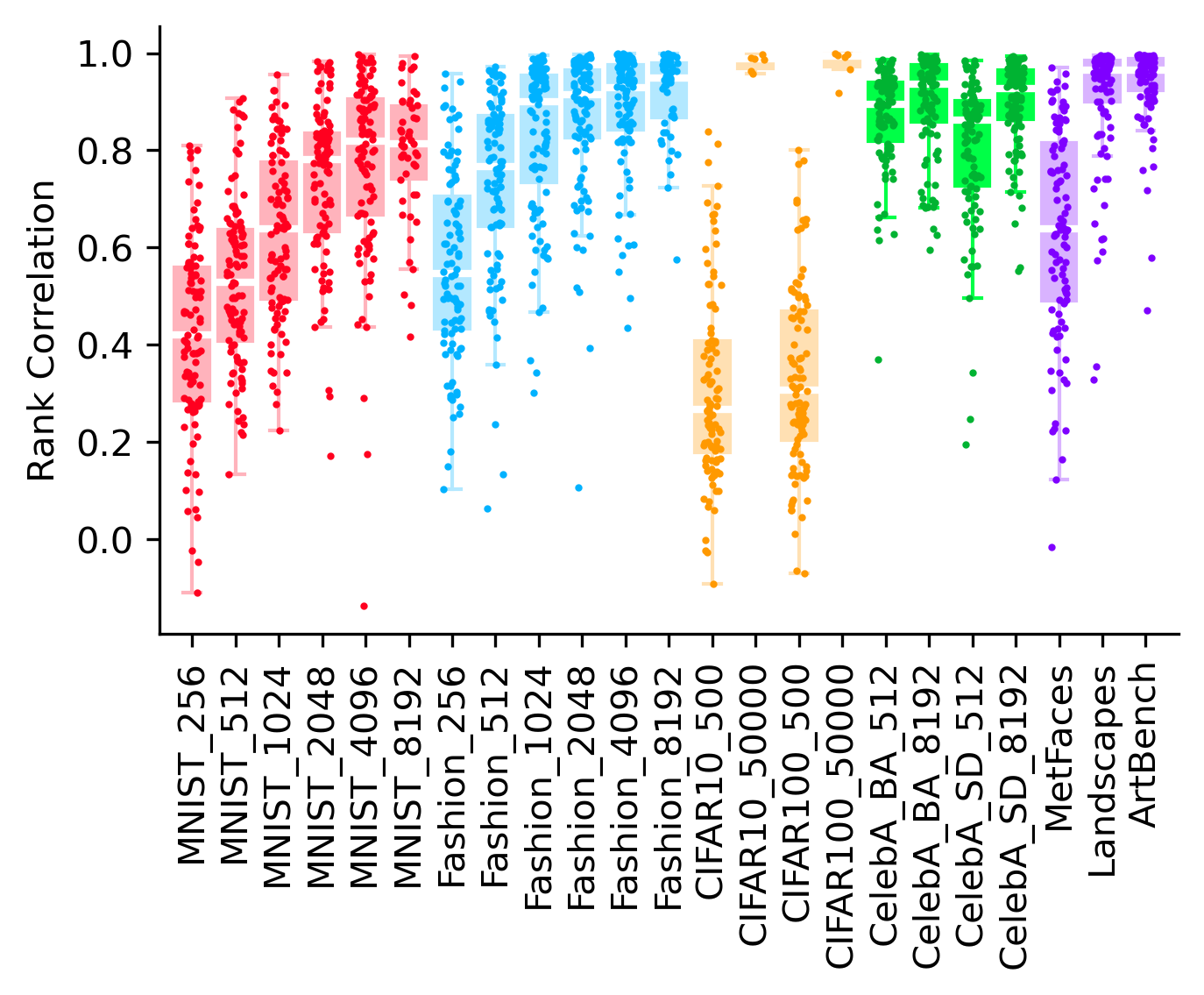}
  \includegraphics[width=0.8\linewidth]{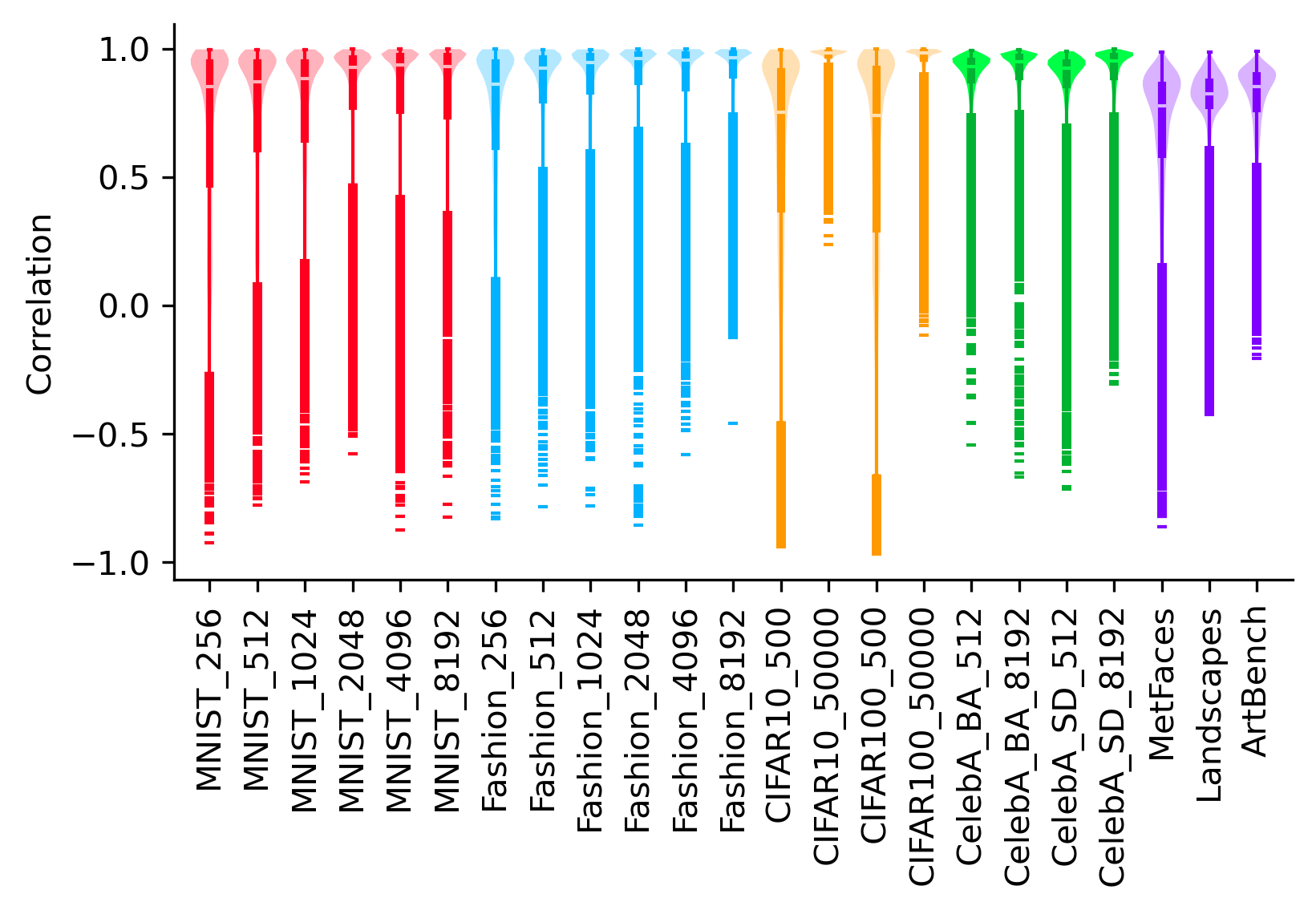}
  \end{center}
  \caption{Differential ablation accurately captures the effect of ablation}
  Top: For each counterfacual landscape we calculated the rank correlation between the Euclidean distances from the factual sample of the counterfactual samples and those distances approximated by differential ablation. The distribution of those rank correlations are given in the top panel as strip plots, with a bar plot in the back to provide additional statistical info. The break in the bar plot indicates the median.

  Bottom: For each counterfactual sample, we calculate its per pixel difference from its factual sample. We then calculate its per pixel difference that is approximated from differential ablation. We then compute the Pearson correlation of these two sets of differences for each counterfactual sample. The distribution of correlations are given as violin plots in the bottom panel, with bar plots with flier points overlaid to provide additional statistical info. The break in the bar plot indicates the median.
  \label{fig4}
\end{figure}
\fi

\subsection{Comparing ablation to retraining}
\label{appdx-res-benchmark}

We then train ensembles of size 16 on subsets of MNIST and FASHION of size 384, where each image is its own data source. We then trained 385 models: one model on the full subset, and one for each leave-one-out subset. To maximize consistency, the initial parameters and the order and contents of the minibatches were kept consistent for the training of each model, and the removed image was replaced with an all black image.

Since ablation and retraining operate on different models, we compare the methods by checking how well visual attribution aligns with counterfactual attribution in each. Table \ref{table1} provides the number of times the top-8 visually attributed data sources intersected with the top-8 counterfactually attributed data sources. Counterfactual attribution was performed using Euclidean distance, while 5 different metrics for visual attributions are reported (see Appendix \ref{appdx-setup-percmetr} for discussion on the metrics).

\begin{table}[h]
\caption{Top-8 Intersections of visual and counterfactual attributions. These values are all significantly above random. Assuming attributions are random, the mean number of intersections would be 158.73 with a standard deviation 11.58.}
\label{table1}
\vskip 0.15in
\begin{center}
\begin{small}
\begin{tabular}{p{5cm} c c c c c}
\toprule
& Euclidean & LPIPS & OpenCLIP & DINOv2 & DINO Patch \\
\midrule
MNIST Ablation & 272 & 262 & 236 & 225 & 267 \\
MNIST Retrain & 283 & 279 & 273 & 261 & 251 \\
MNIST Differential ablation & 838 & 822 & 715 & 667 & 747 \\
Fashion Ablation & 296 & 272 & 250 & 252 & 264 \\
Fashion Retrain & 306 & 303 & 278 & 284 & 280 \\
Fashion Differential Ablation & 659 & 605 & 550 & 500 & 538 \\
\bottomrule
\end{tabular}
\end{small}
\end{center}
\vskip -0.1in
\end{table}

We see with all metrics, ablation and retraining appears to be similar, with retraining being a little more closely aligned to visual attribution. Differential ablation is significantly more aligned to visual attribution than either of the two.

We can also observe this by computing how visually similar the counterfactually attributed training images are to the factual image. We take the top counterfactually attributed image, and calculate its visual similarity based on one of the 5 perceptual metrics. We then calculate its rank among all images in the training set and normalize the rank to be between 0 and 1, so that the visually closest image is assigned 0 and the visually furthest image is assigned 1. The distributions of visual similarities of the top counterfactually attributed images are presented in Figure \ref{figSupBen1}. Again, ablation and retraining are comparable, while differential ablation attributes images that are much more visually similar.

\ifrenderfigures
\begin{figure}
  \begin{center}
  \includegraphics[width=0.9\linewidth]{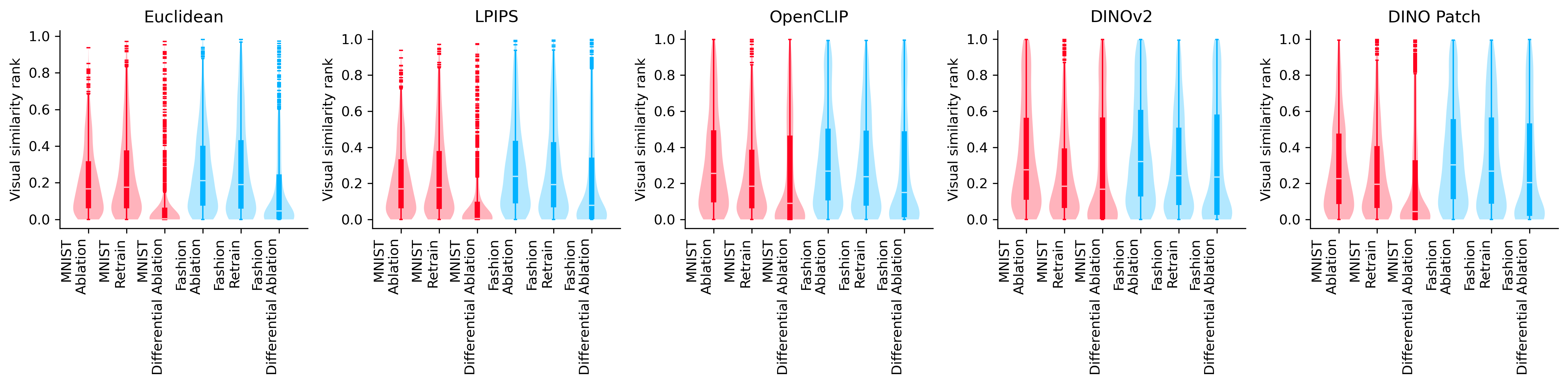}
  \end{center}
  \caption{Visual similarity between factual image and top counterfactually attributed image}
  \label{figSupBen1}
  Distributions of visual similarty are given in violin plots, which are overlaid with box plots to provide additional statistical info.
\end{figure}
\fi

Counterfactual attribution can also be performed with different perceptual metrics, as described in Section \ref{sec-res-benchmark}. We compare the top-1, top-3, and top-8 intersections for all visual and counterfactual attribution methods in Figure \ref{figSupBen} (for differential attribution we only use Euclidean distances). We note that the agreement between the various different counterfactual methods is less than that in visual attribution. This may be due to better alignment of the metrics at smaller distances: visual attribution attempts to find the closest image, while counterfactual attribution attempts to find the most distant counterfactual. In all cases we can observe that differential ablation is more closely aligned with visual attribution than counterfactual attribution.

\ifrenderfigures
\begin{figure}
  \begin{center}
  \includegraphics[width=0.48\linewidth]{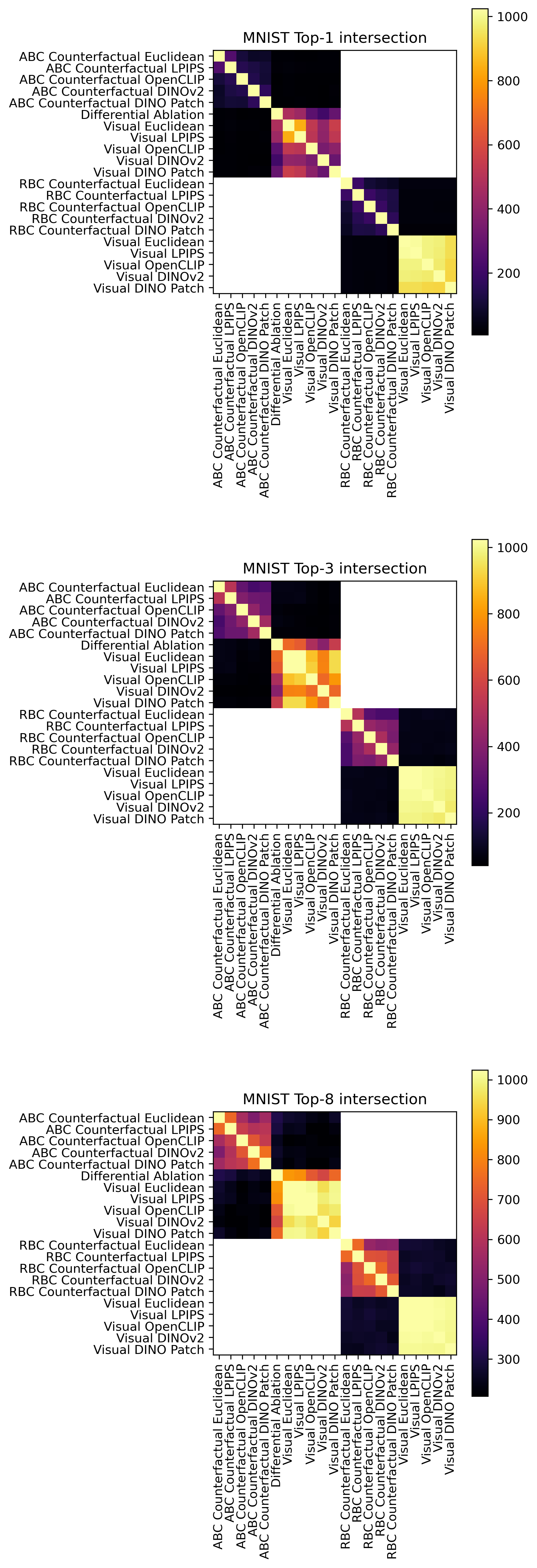}
  \includegraphics[width=0.48\linewidth]{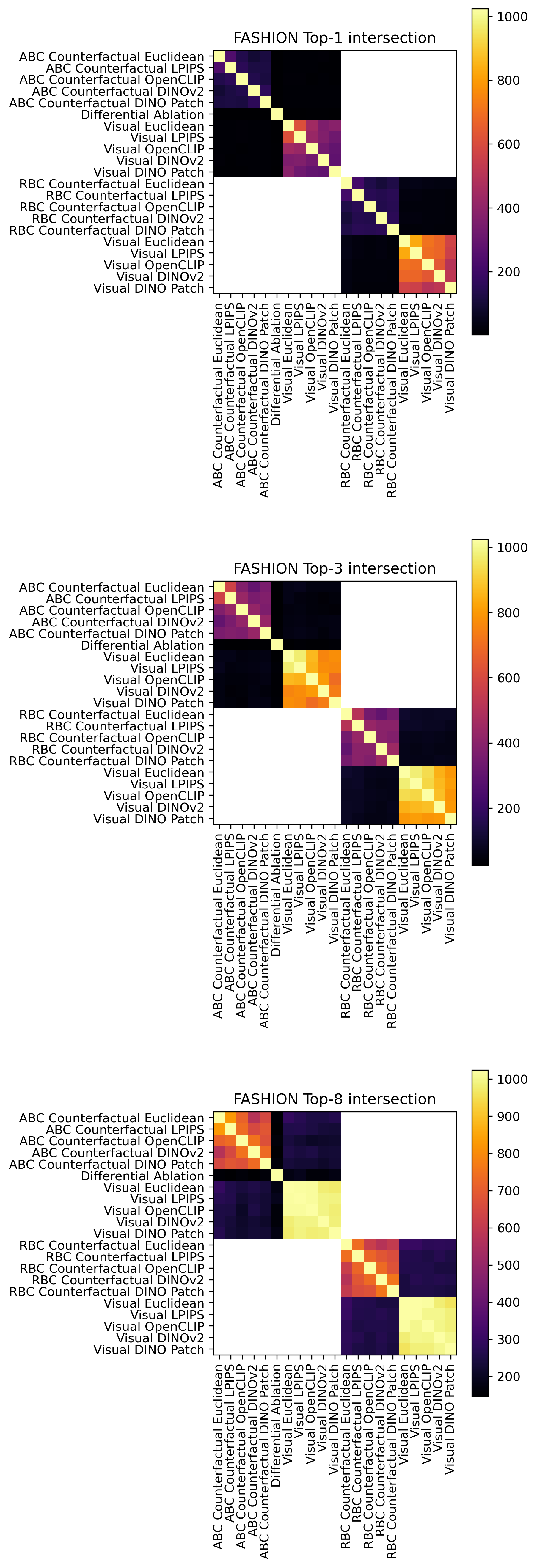}
  \end{center}
  \caption{Top-$n$ intersections between various visual and counterfactual attributions}
  \label{figSupBen}
\end{figure}
\fi

\subsection{Visual similarity of images attributed with differential ablation}
\label{appdx-res-diffablres}

We use differential ablation to counterfactually attribute 1000 generated samples to data sources for each of our 23 ensembles. We then calculate the visual similarity rank of the attributed data sources. This is calculated by taking each data source and calculating the minimum distance between training points generated by it and the generated sample. The data sources are then ranked, and the rank is normalized to between 0 and 1, so that the data source with a 0 is the most visually similar to the generated sample, and the data source with a 1 is the most visually distant. Different visual similarity ranks can be calculated by adjusting how the distance is measured. The distributions of visual similarity ranks of the counterfactually attributed samples are given in Figure \ref{fig5} (for the Euclidean distance) and Figure \ref{figSupDAVS} for 4 other perceptual metrics (LPIPS, OpenCLIP, DINOv2, DINO Patch, see Appendix \ref{appdx-setup-percmetr} for discussion on the metrics).

\ifrenderfigures
\begin{figure}
  \begin{center}
  \includegraphics[width=0.48\linewidth]{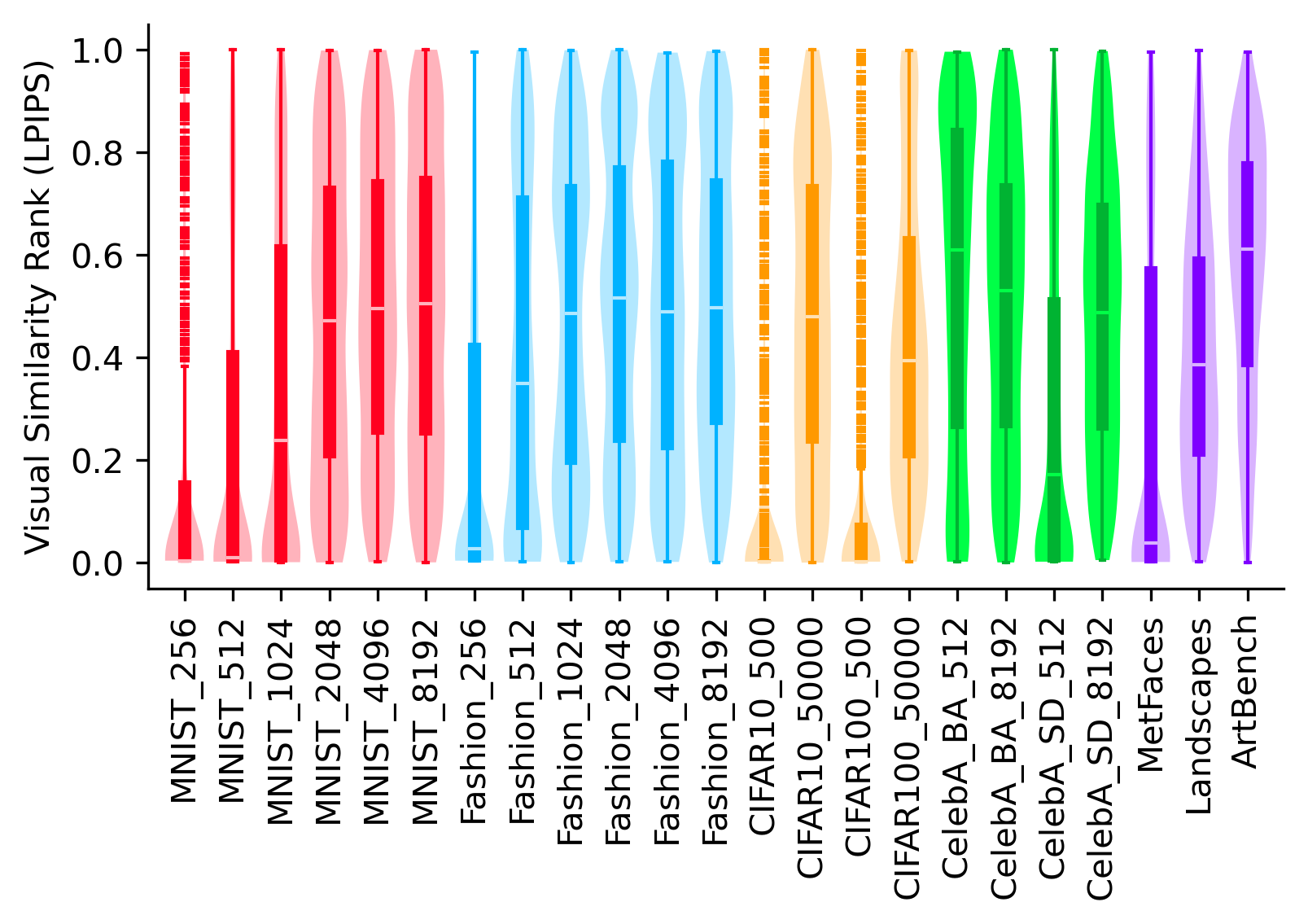}
  \includegraphics[width=0.48\linewidth]{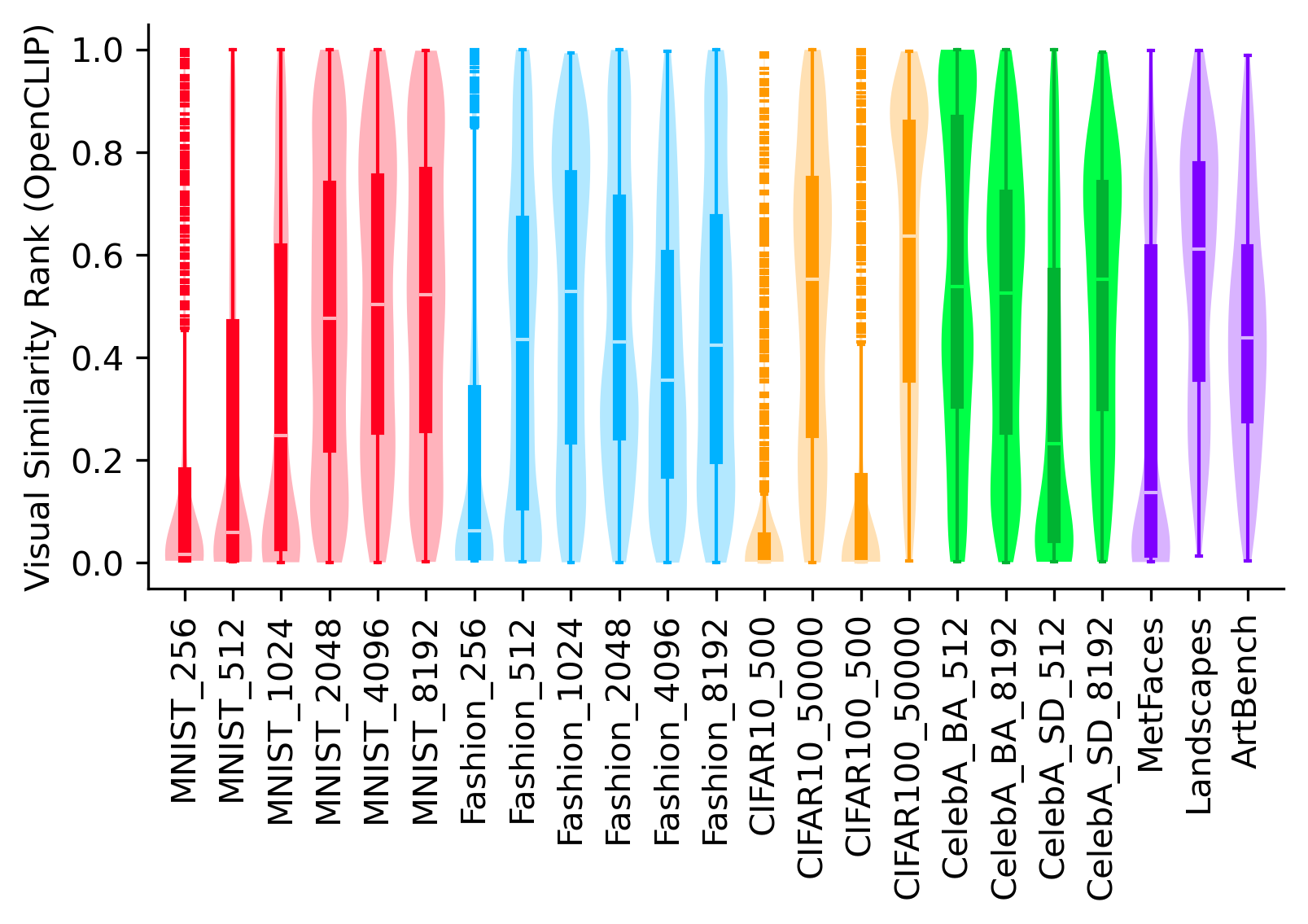}
  
  \includegraphics[width=0.48\linewidth]{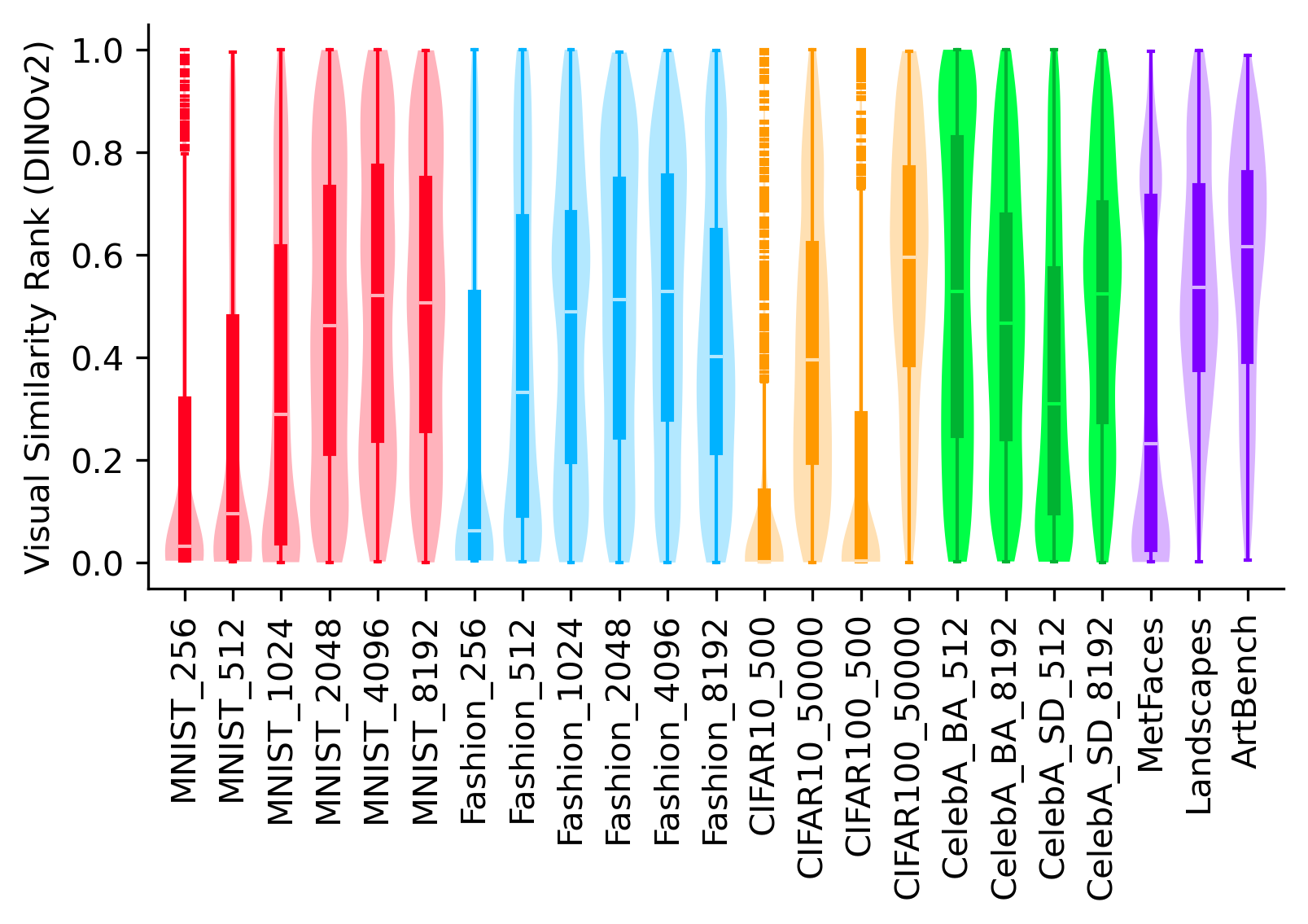}
  \includegraphics[width=0.48\linewidth]{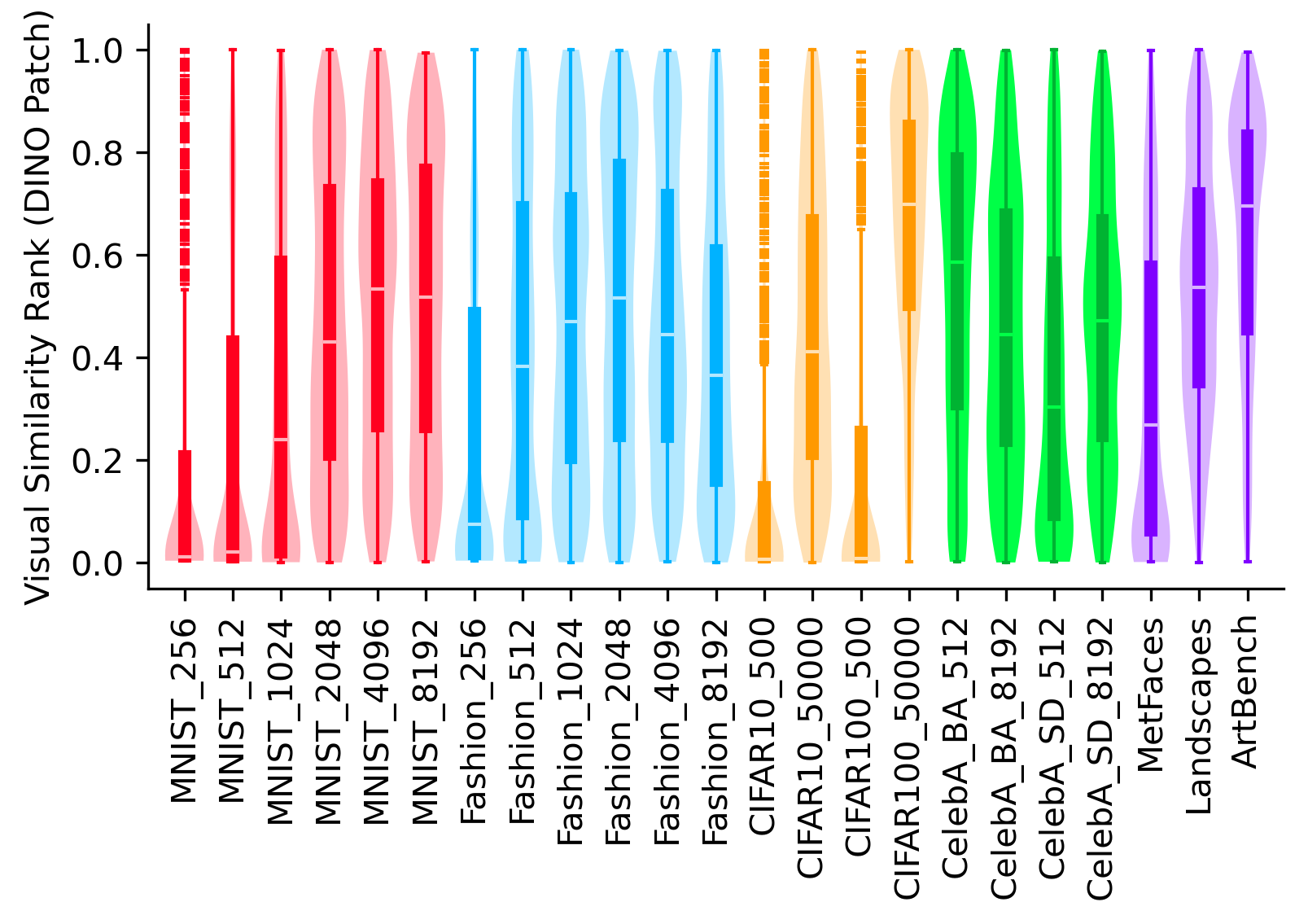}
  \end{center}
  \caption{Distributions of visual similarity ranks (computed using various perceptual metrics) for counterfactually attributed data sources.}
  \label{figSupDAVS}
\end{figure}
\fi

We also randomly select 18 of the 1000 generated samples from each ensemble, and present them next to their attributed data sources in Appendix \ref{appdx-largefig-1}. If a data source is responsible for multiple images, then the image representing the data source is the image originating from the data source that is the closest to the attributed image in terms of flip agnostic Euclidean distance, which we define to be the minimum of the Euclidean distance between two images and the Euclidean distance between the same two images, but one of them is horizontally flipped.

\subsection{Divergence in visual and counterfactual attribution}
\label{appdx-res-cdr}

For each of the 23 ensembles, we generated 100 samples and their full counterfactual landscapes, with the exception of MNIST\_8192 and FASHION\_8192 where we generated 50, and CIFAR-10\_50000 and CIFAR-100\_50000 where we generated 6. For each generated sample, we visually attributed them to the training set with one of 5 different perceptual metrics (Euclidean, LPIPS, OpenCLIP, DINOv2, DINO Patch, see Appendix \ref{appdx-setup-percmetr} for discussion on the metrics). We then calculate the counterfactual distance rank associated with the top visually attributed training sample. This is calculated by enumerating the entire counterfactual landscape of the generated sample. The counterfactuals are then ranked by their Euclidean distance to the generated sample and scaled the ranks so all values are between 0 and 1, with 0 being close to the factual sample and 1 being far. The score reported is then the value assigned to the counterfactual whose data source produced the training datum that was visually attributed.

The distributions of the counterfactual distance ranks of the visually attributed samples are provided in Figure \ref{fig6} (for visual attributions that are made using the Euclidean distance) and Figure \ref{figSupCFVIS} (for other visual attributions made using LPIPS, OpenCLIP, DINOv2, and DINO Patch). See Appendix \ref{appdx-setup-percmetr} for discussion on the different metrics.

\ifrenderfigures
\begin{figure}
  \begin{center}
  \includegraphics[width=0.48\linewidth]{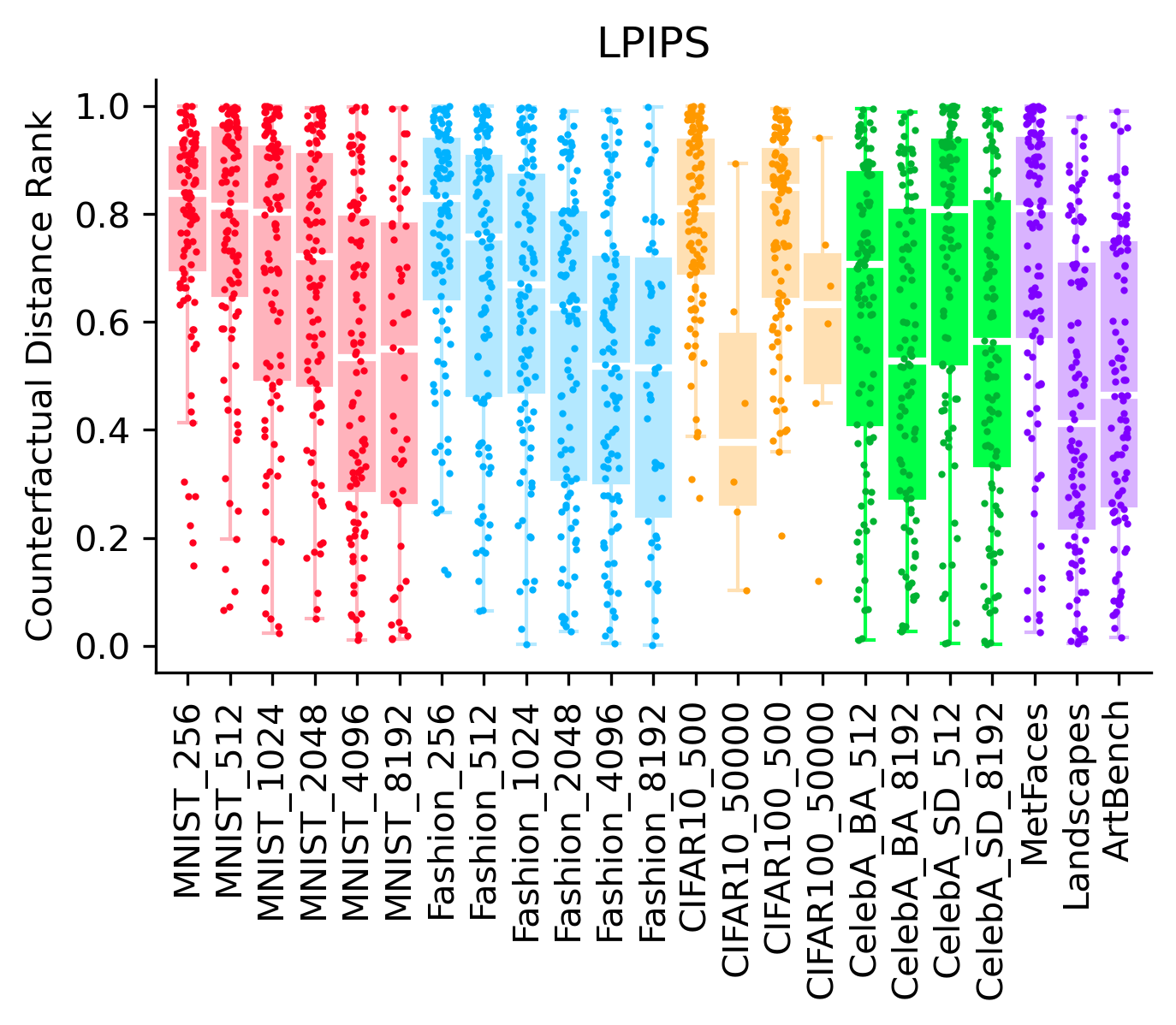}
  \includegraphics[width=0.48\linewidth]{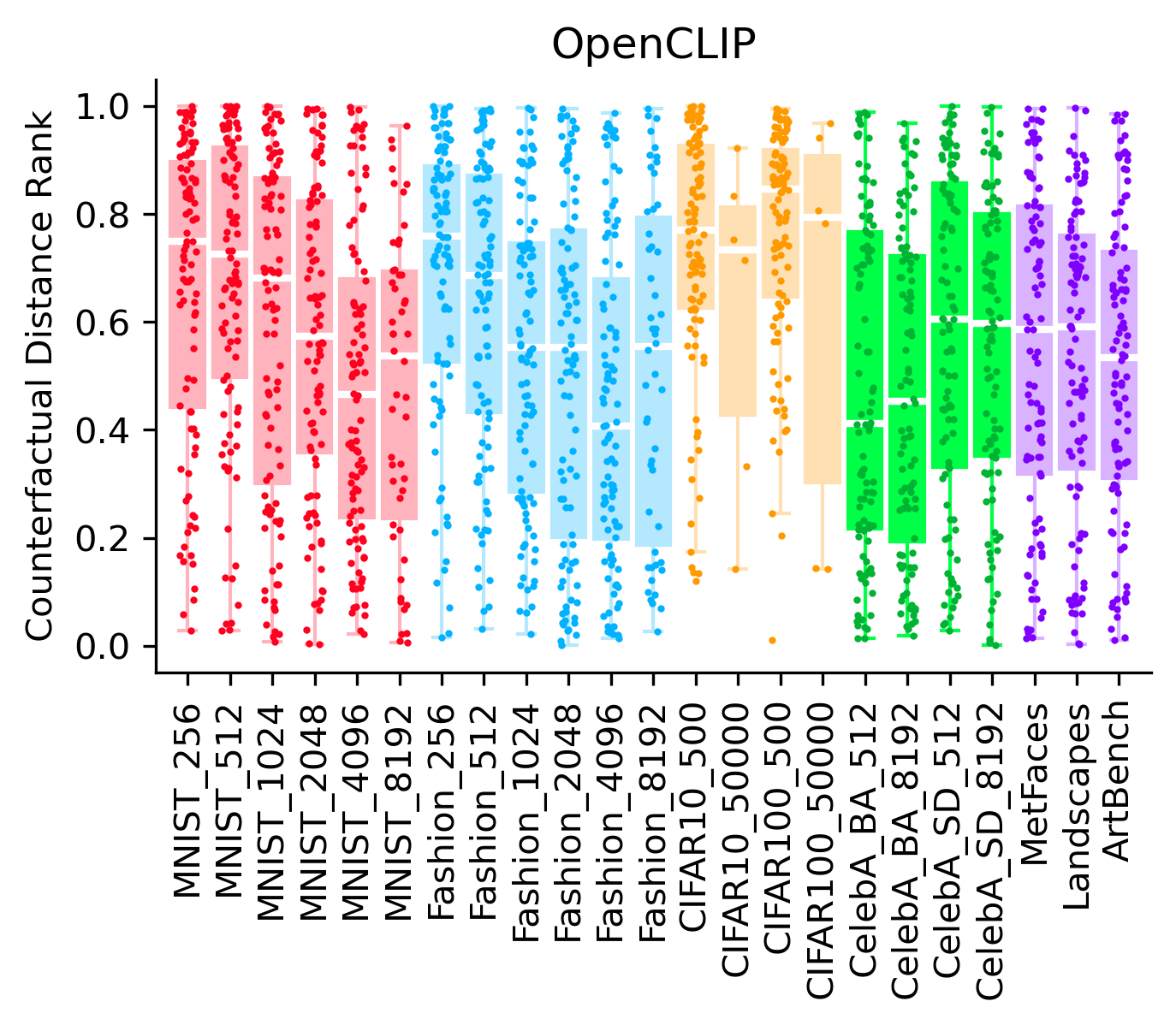}
  
  \includegraphics[width=0.48\linewidth]{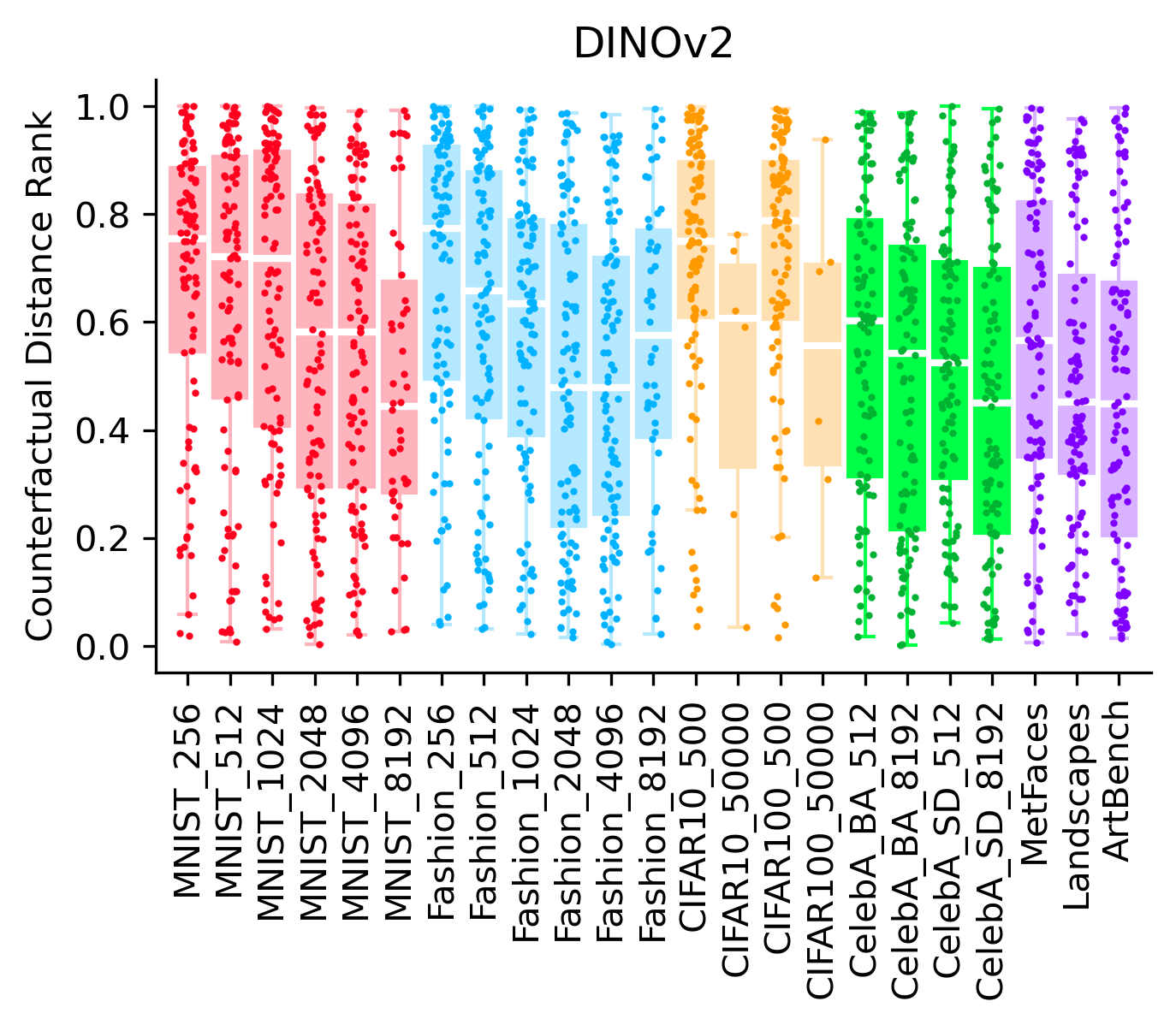}
  \includegraphics[width=0.48\linewidth]{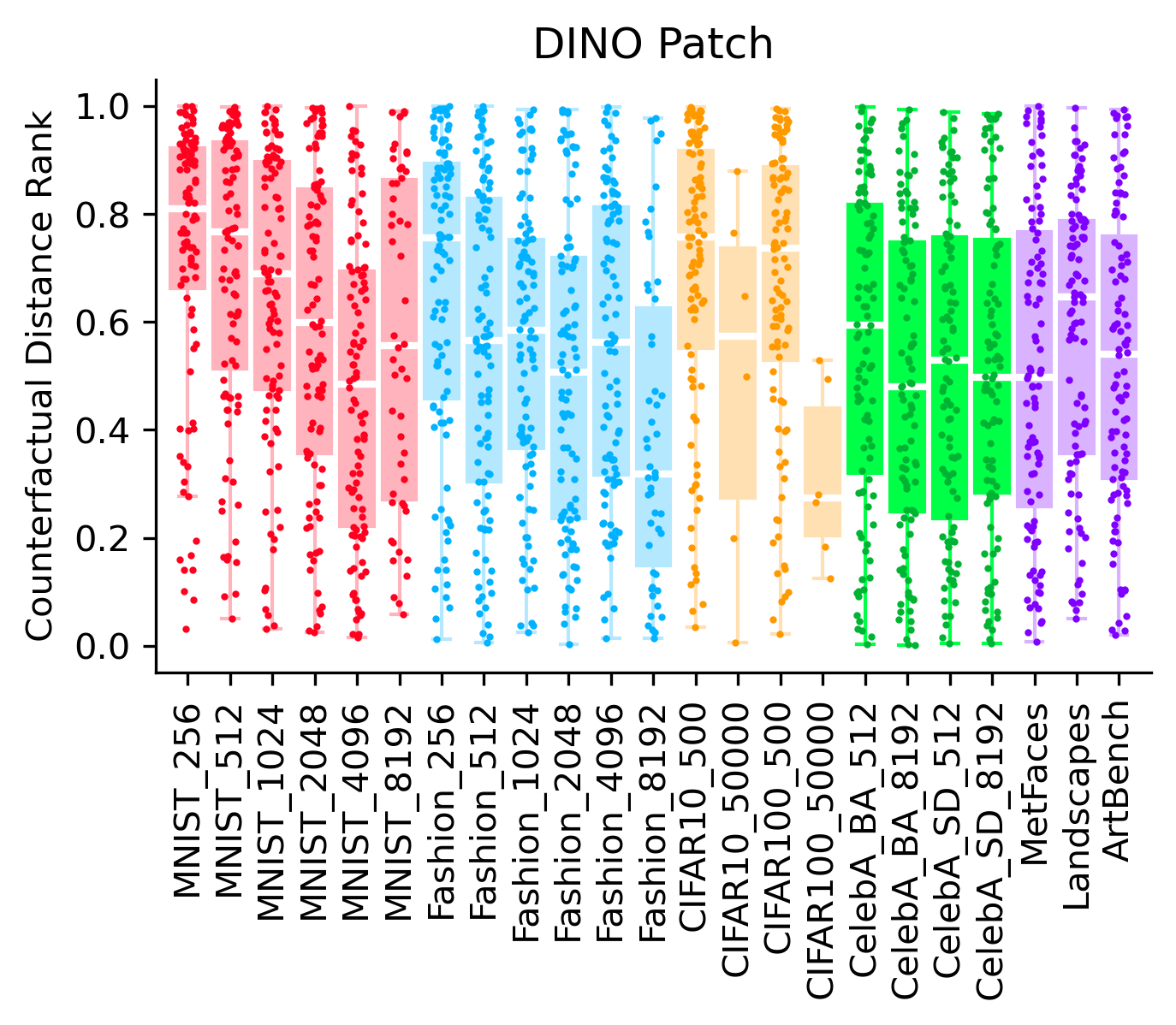}
  \end{center}
  \caption{Distributions of counterfactual distance ranks for data sources that were visually attributed using LPIPS, OpenCLIP, DINOv2, and DINO Patch as perceptual metrics.}
  \label{figSupCFVIS}
\end{figure}
\fi

We also randomly select 18 of the generated samples from each ensemble (except CIFAR-10\_50000 and CIFAR-100\_50000, where only 6 samples were generated), and present them next to their attributed training data in Appendix \ref{appdx-largefig-2}. We also present the counterfactual sample where that piece of training data is removed.

\subsection{Counterfactual radii}
\label{appdx-res-cfr}

We calculate the counterfactual radii of various generated samples measured in non-Euclidean perceptual metrics (see Appendix \ref{appdx-setup-percmetr} for discussion on the metrics), and present their distributions in Figure \ref{figSupCFR}. The distribution of counterfactual radii measured in Euclidean distance can be found in Figure \ref{fig7}.

\ifrenderfigures
\begin{figure}
  \begin{center}
  \includegraphics[width=0.48\linewidth]{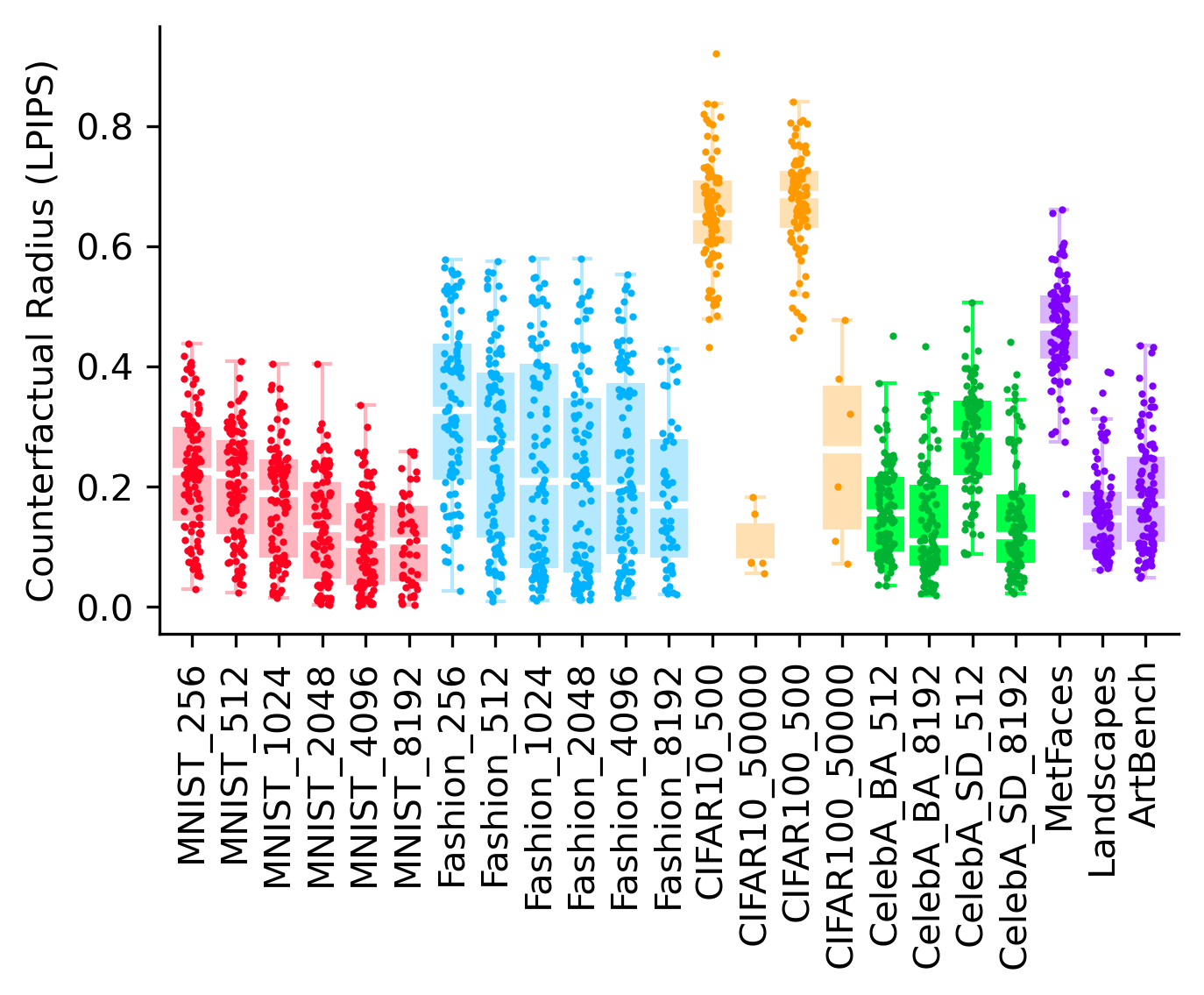}
  \includegraphics[width=0.48\linewidth]{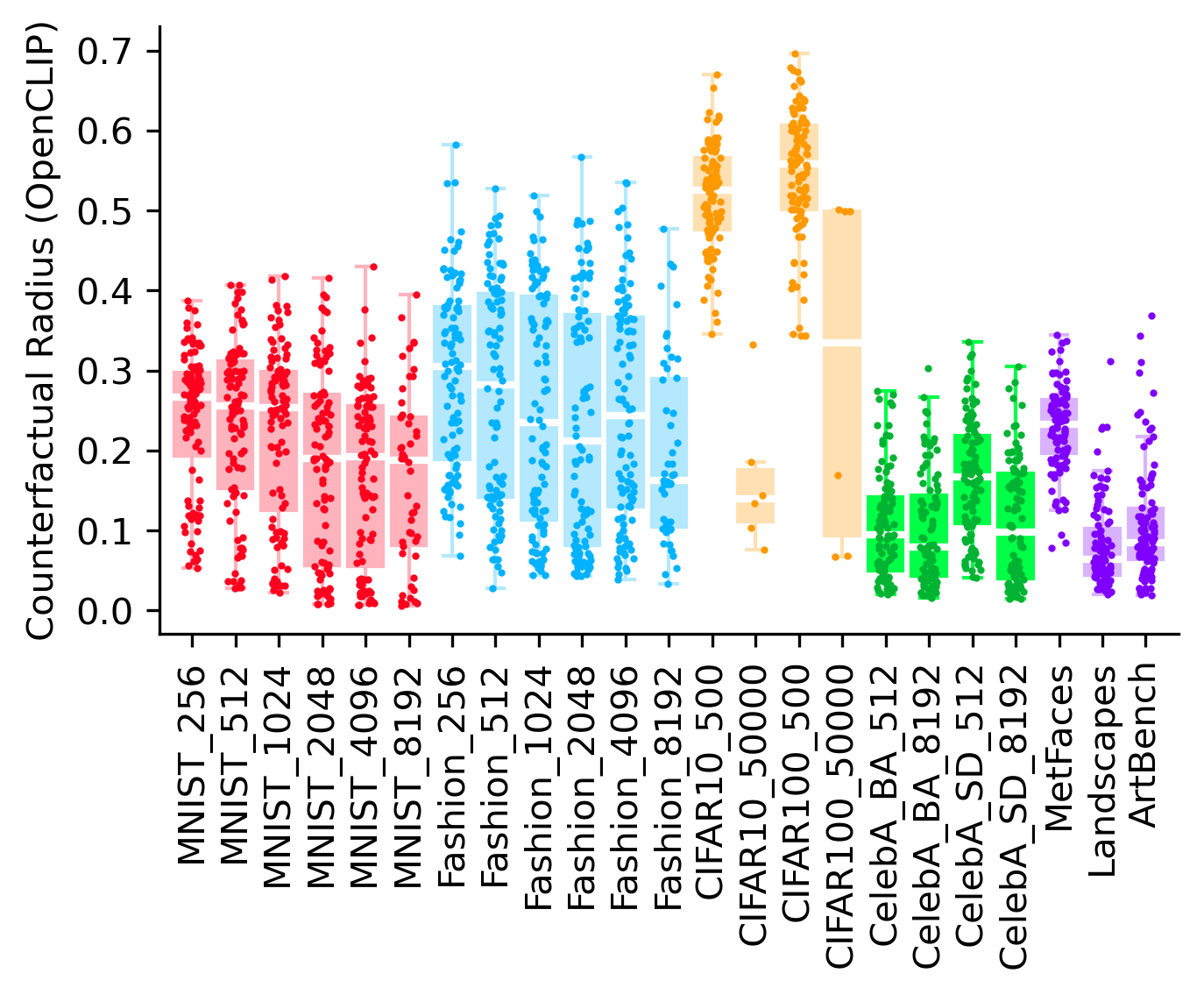}
  
  \includegraphics[width=0.48\linewidth]{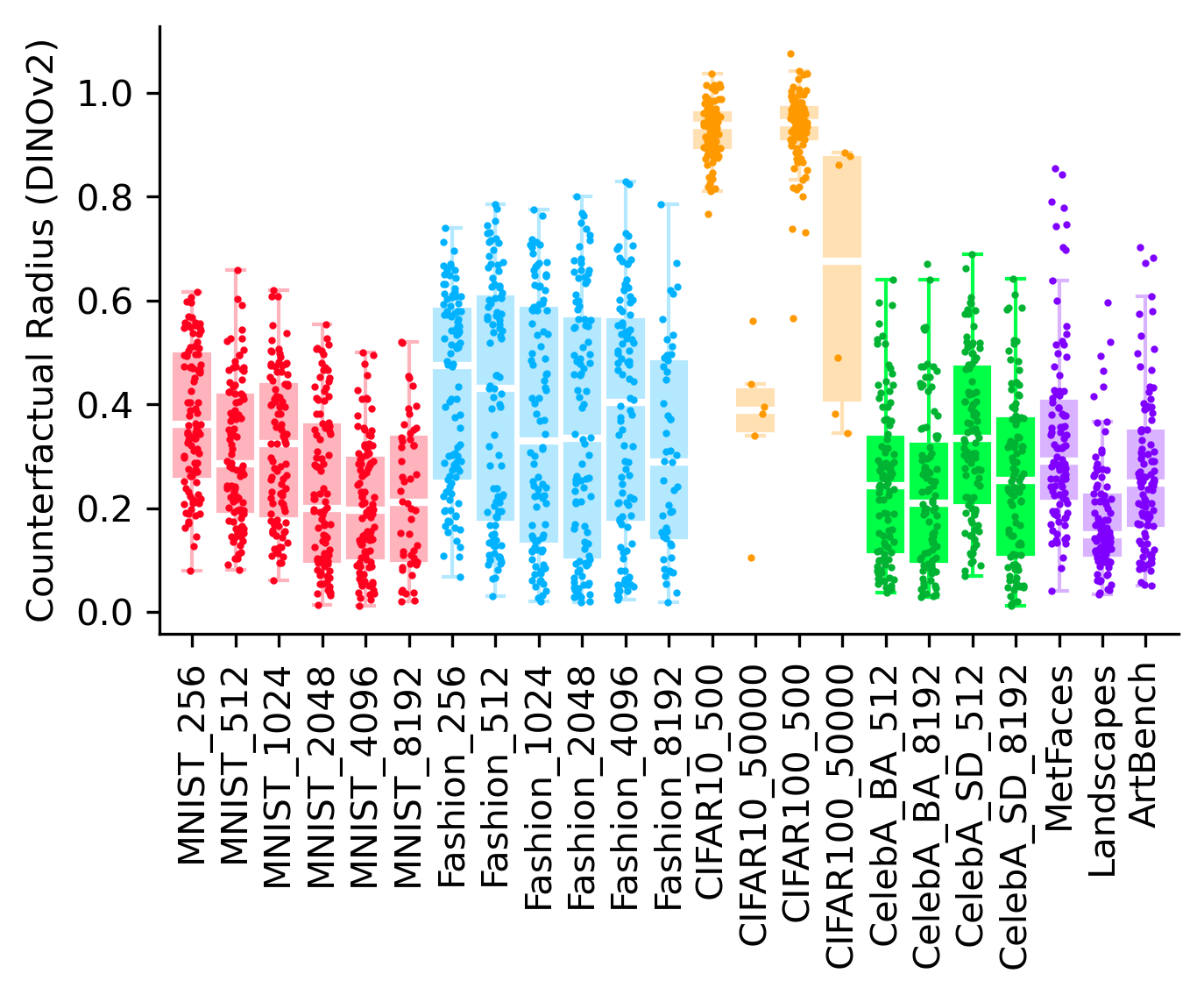}
  \includegraphics[width=0.48\linewidth]{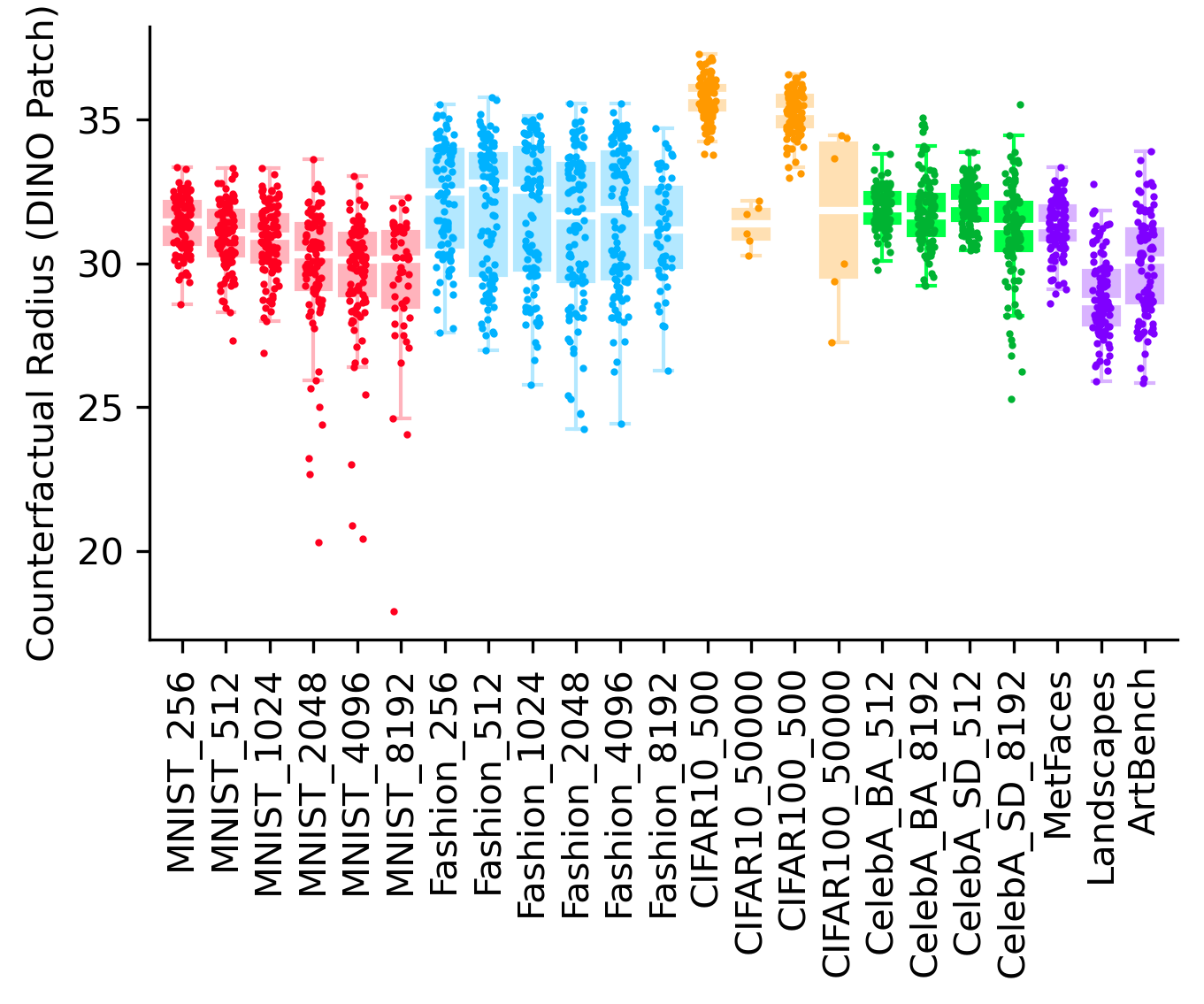}
  \end{center}
  \caption{Distributions of counterfactual radii generated from various ensembles that are measured in LPIPS, OpenCLIP, DINOv2, and DINO Patch perceptual metrics.}
  \label{figSupCFR}
\end{figure}
\fi

We also plot the relation between the counterfactual radii and training set size in 
Figure \ref{figsupradrel}. We fit a line of best fit via linear regression and report the line of best fit along with the $R^2$ value. The p-values for a zero slope null hypothesis are 0.02 for LPIPS, 0.0002 for OpenCLIP, 0.06 for DINOv2, and 0.14 for DINO Patch.

\ifrenderfigures
\begin{figure}
  \begin{center}
  \includegraphics[width=0.8\linewidth]{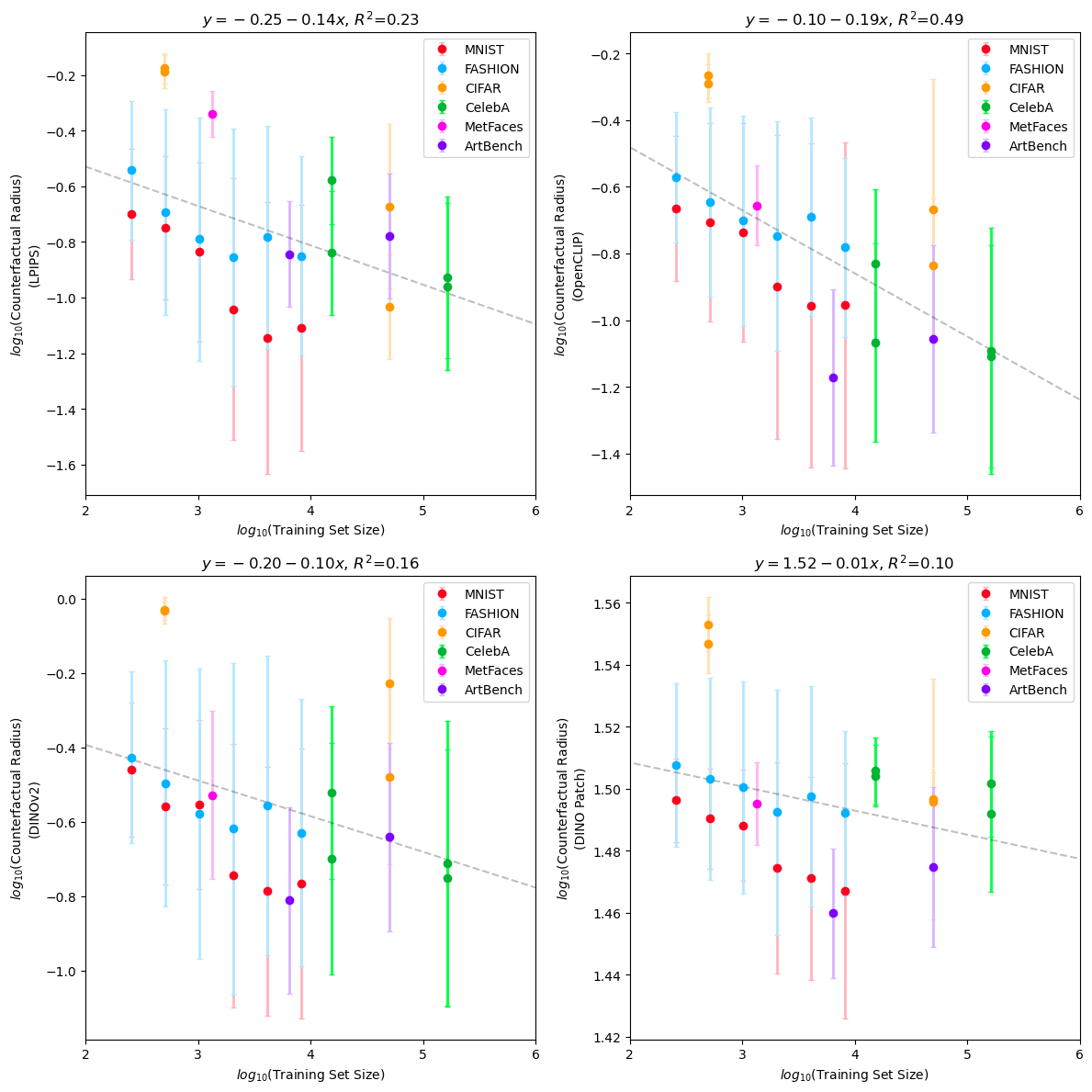}
  \end{center}
  \caption{Relation between counterfactual radii (measured in LPIPS, OpenCLIP, DINOv2, and DINO Patch) and training data size}
  For each ensemble, we compute the counterfactual radii of samples generated from it. We then take the base 10 log of each radius. The mean of these distributions are indicated by the y value of the points, and the error bars extend for one standard deviation of the distribution.
  \label{figsupradrel}
\end{figure}
\fi
\clearpage
\section{Large figures}

Large multi-page figures are confined to this section to avoid breaking the flow of the rest of the paper.

\ifrenderfigures
\subsection{Generated samples and their counterfactually attributed data sources via differential ablation}
\label{appdx-largefig-1}

\begin{figure}[h]
  \centering
  \includegraphics[width=\linewidth]{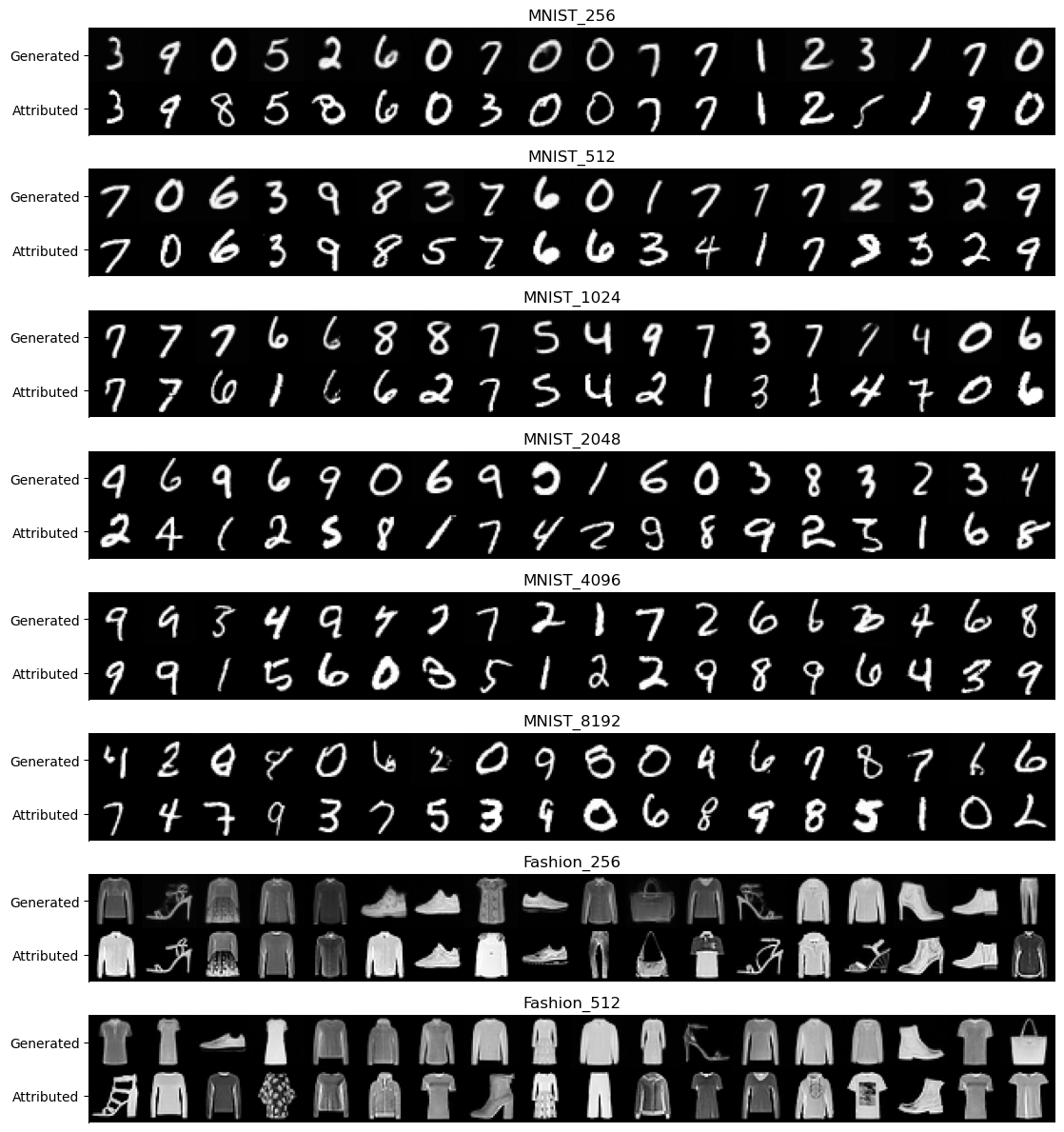}
\end{figure}
\begin{figure}[b]
  \centering
  \includegraphics[width=\linewidth]{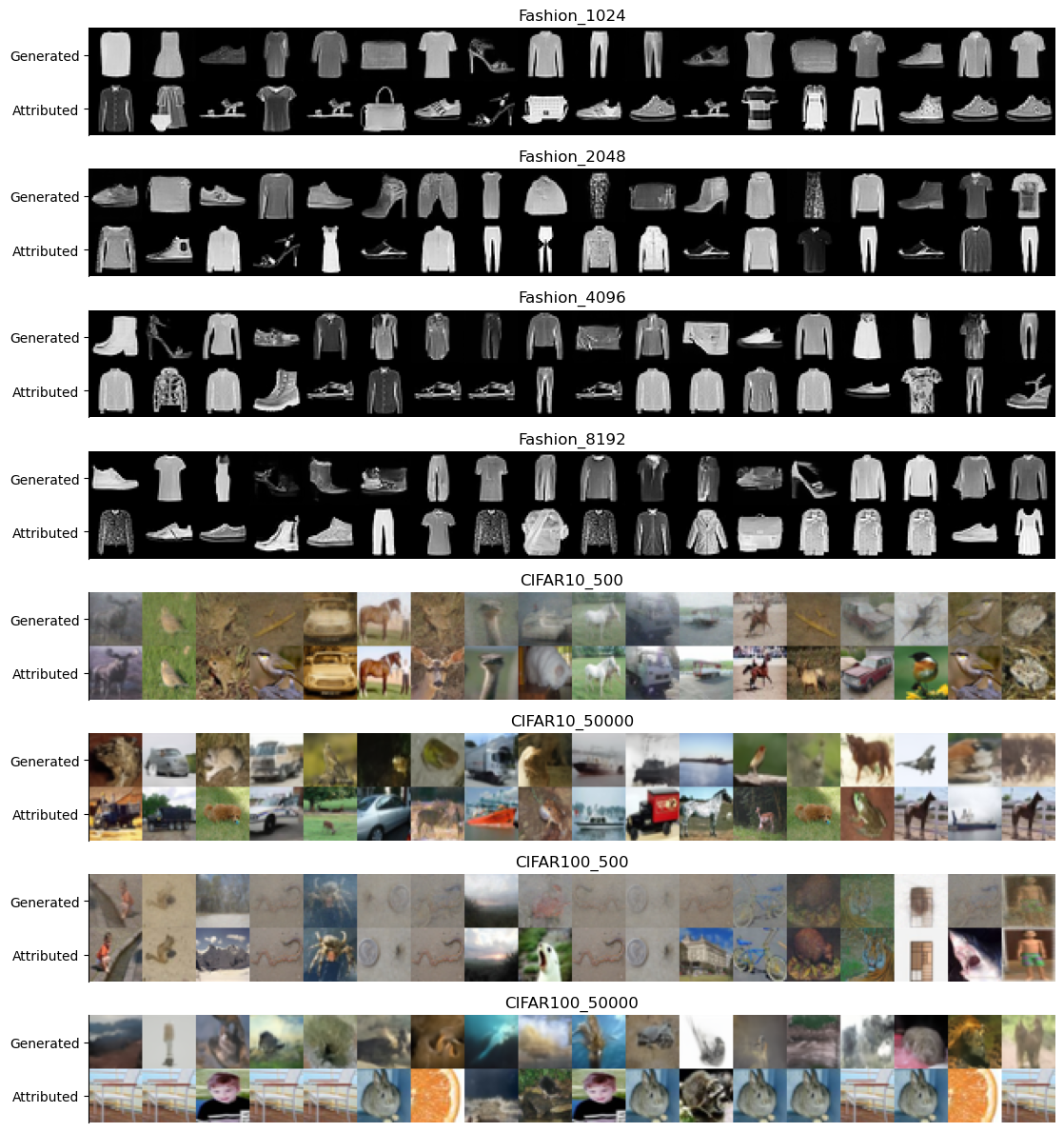}
\end{figure}
\begin{figure}[b]
  \centering
  \includegraphics[width=\linewidth]{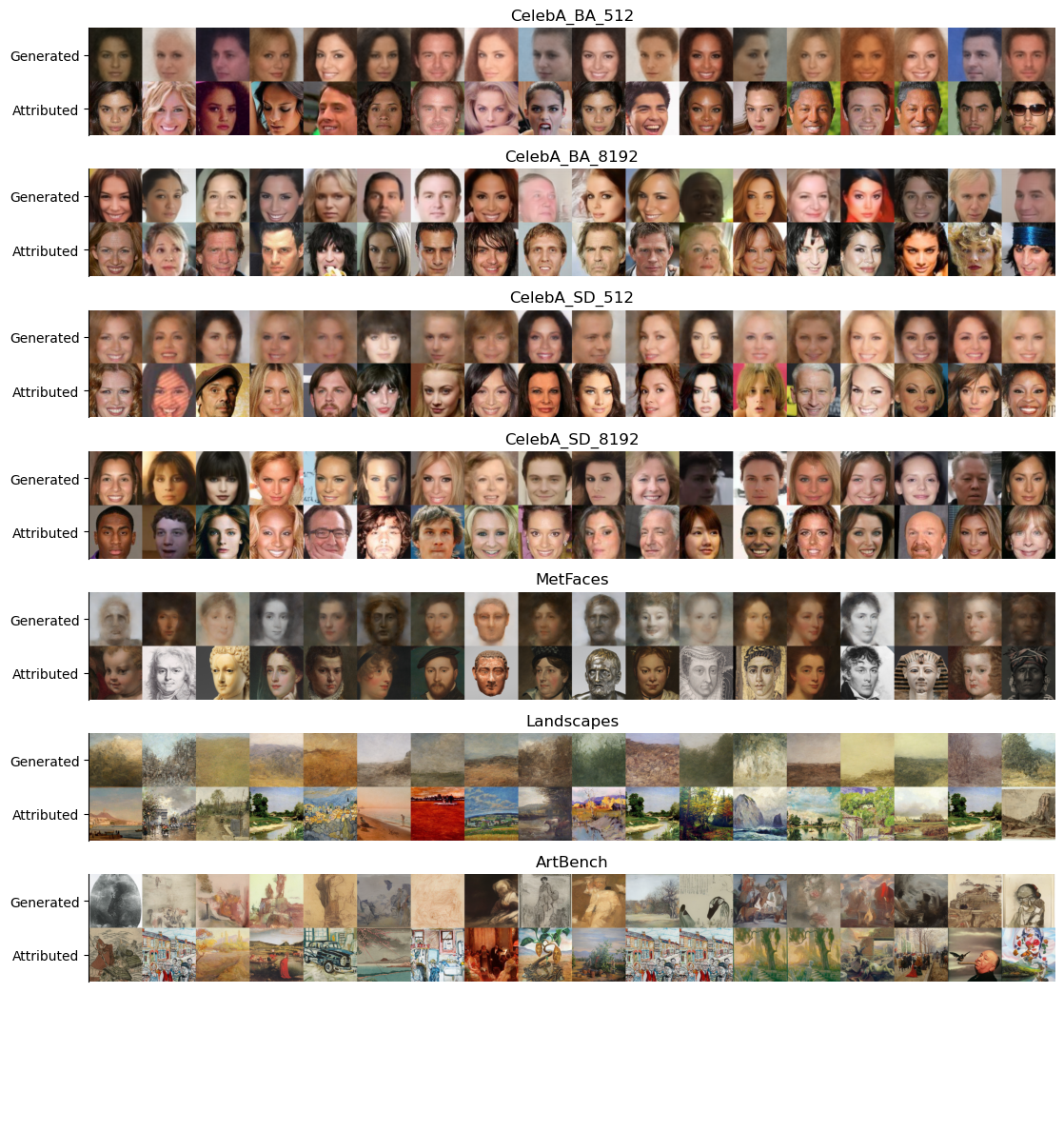}
\end{figure}

\clearpage

\subsection{Generated samples and their visually attributed training samples and counterfactuals samples where the visually attributed sample is removed}
\label{appdx-largefig-2}
\begin{figure}[h]
  \centering
  \includegraphics[width=0.9\linewidth]{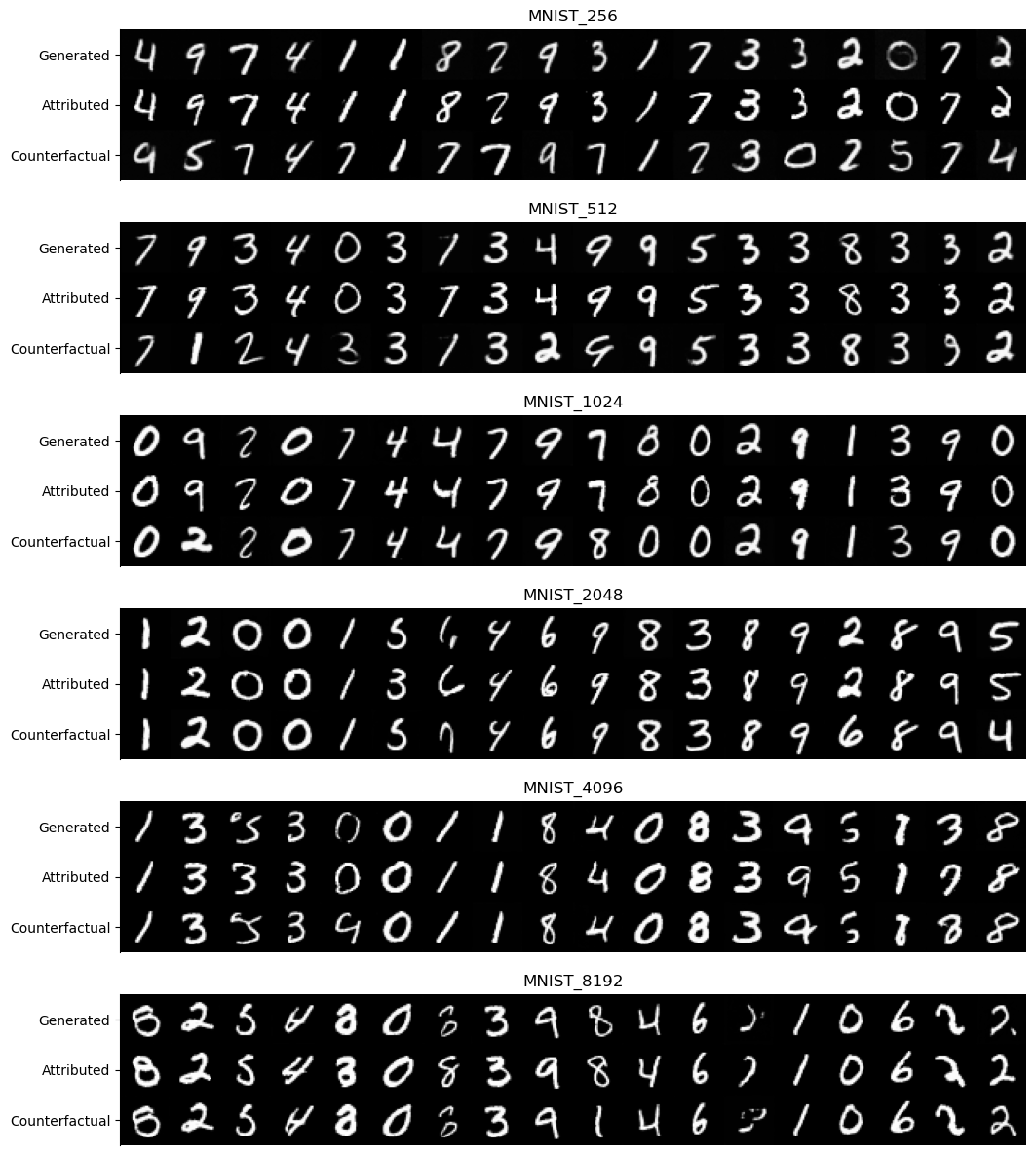}
\end{figure}

\begin{figure}[b]
  \centering
  \includegraphics[width=\linewidth]{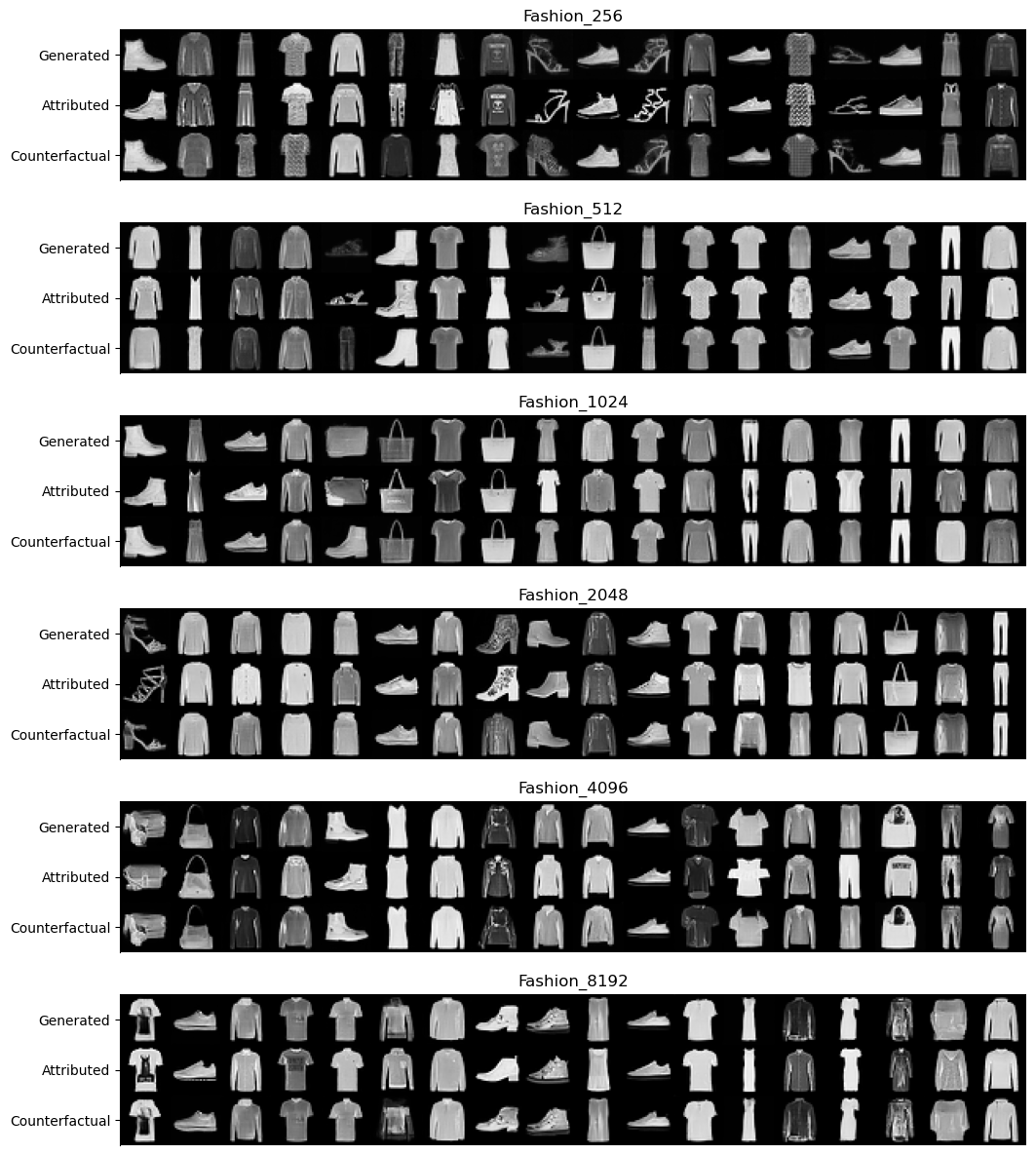}
\end{figure}

\begin{figure}[b]
  \centering
  \includegraphics[width=\linewidth]{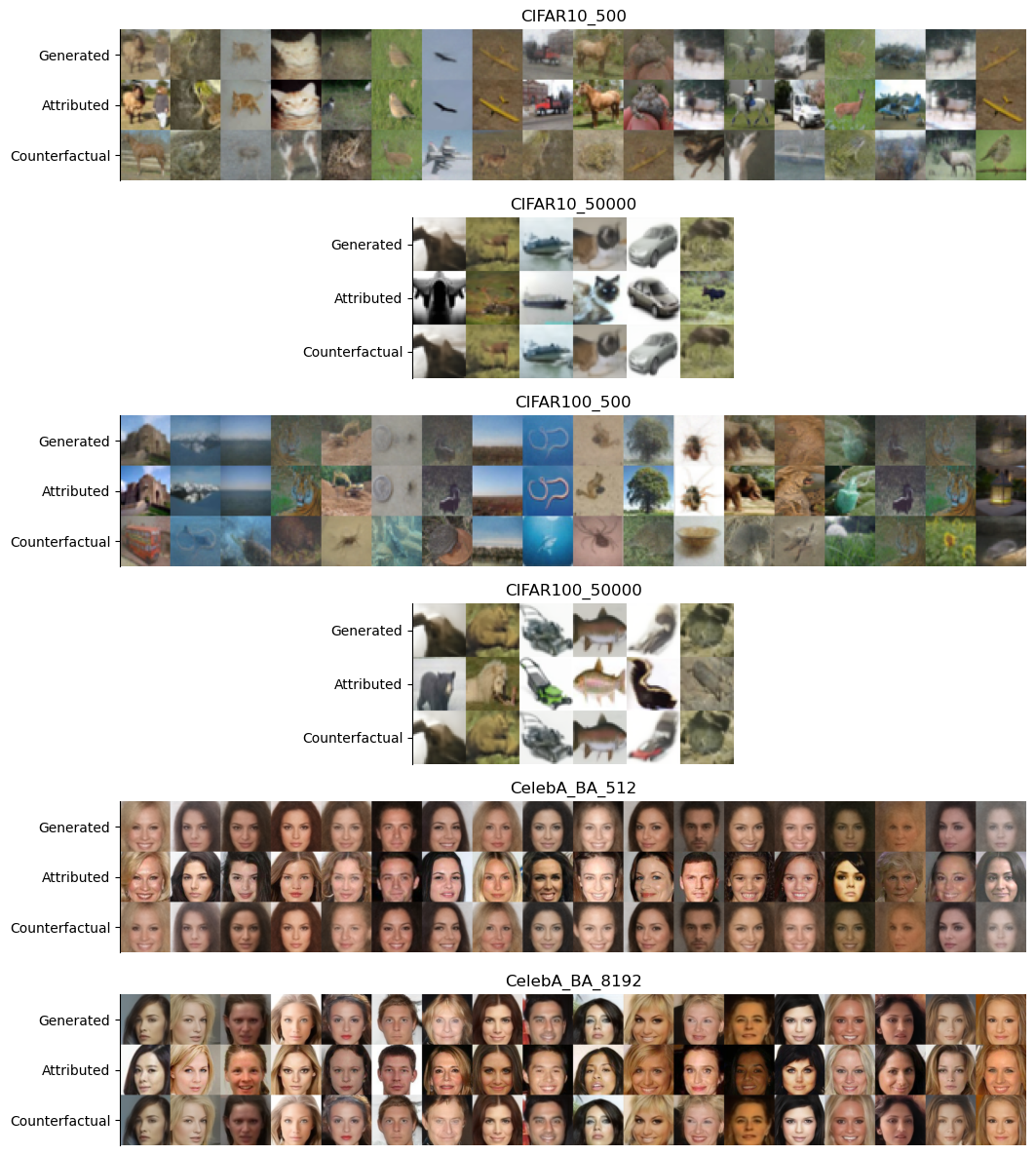}
\end{figure}

\begin{figure}[b]
  \centering
  \includegraphics[width=\linewidth]{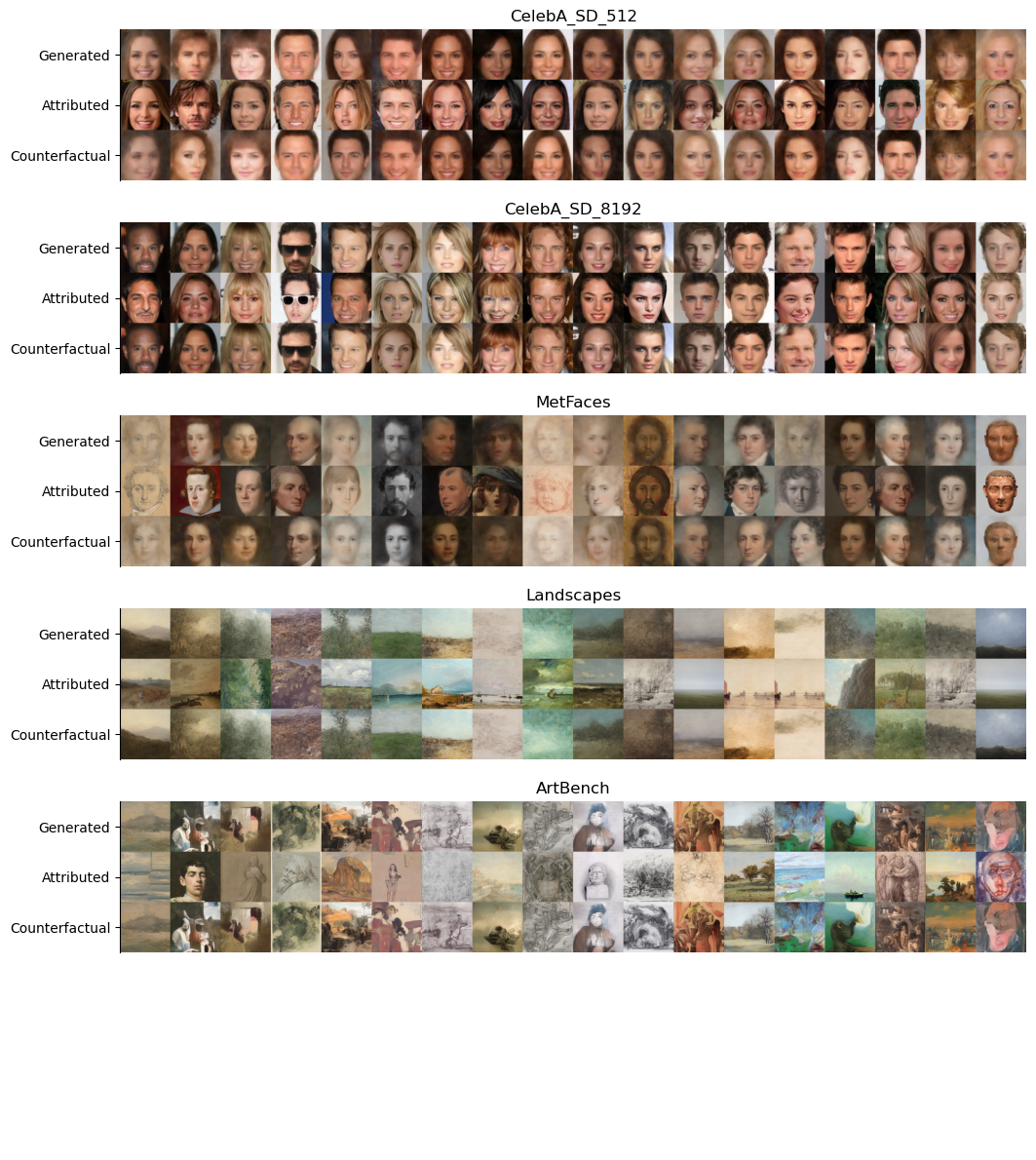}
\end{figure}

\clearpage

\subsection{Entire counterfactual landscape of a generated MNIST digit and a generated CelebA face}
\label{appdx-largefig-3}

\begin{figure}[h]
  \centering
  \includegraphics[width=0.5\linewidth]{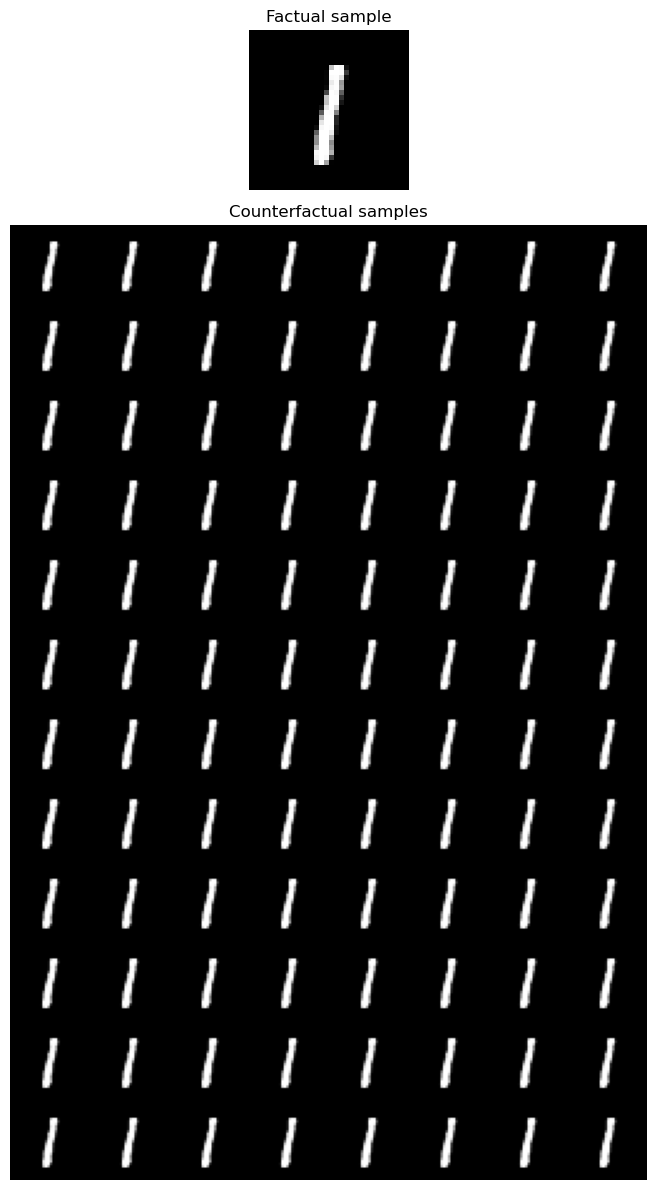}
\end{figure}

\begin{figure}[b]
  \centering
  \includegraphics[width=0.7\linewidth]{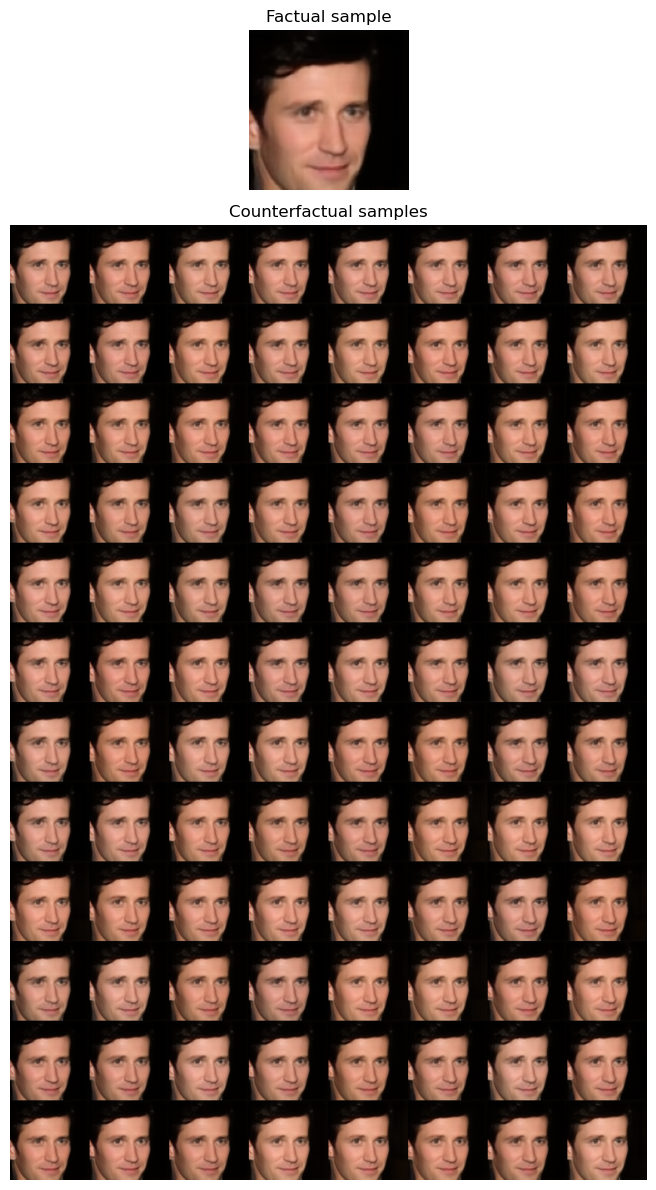}
\end{figure}
\fi

%
%
%
\bibliography{references}
\end{document}